\documentclass[10pt,journal,compsoc]{IEEEtran}

\usepackage{times}
\usepackage{epsfig}
\usepackage{graphicx}
\usepackage{amsmath}
\usepackage{amssymb}
\usepackage{multirow}
\usepackage{subfigure}
\usepackage{color}
\usepackage{rotating}
\usepackage{array}
\usepackage{booktabs}
\usepackage{mdwlist}
\usepackage{float}
\usepackage{url}
\usepackage{cite}
\usepackage{amsfonts}
\usepackage{threeparttable}
\usepackage{enumerate}
\usepackage{ragged2e}
\usepackage[noend]{algpseudocode}
\usepackage{algorithmicx,algorithm}
\usepackage{pifont}         

\newcommand{\cmark}{\ding{51}}
\newcommand{\xmark}{\ding{55}}

\usepackage{caption}
\newlength\savewidth\newcommand\shline{\noalign{\global\savewidth\arrayrulewidth
  \global\arrayrulewidth 1pt}\hline\noalign{\global\arrayrulewidth\savewidth}}

\usepackage[pagebackref=true,breaklinks=true,colorlinks,bookmarks=false]{hyperref}
\ifCLASSINFOpdf
\else
\fi
\begin{document}




\renewcommand{\baselinestretch}{0.998}
\renewcommand{\justify}{\leftskip=0pt \rightskip=0pt plus 0cm}



%
\title{Hierarchical Augmentation and Distillation for Class Incremental Audio-Visual Video Recognition}
%
%
%
%

\author{${\text{Yukun Zuo}}^*$,~\IEEEmembership{Member,~IEEE},\thanks{Yukun Zuo and Hantao Yao contribute equally to the this work.}
        $\text{Hantao~Yao}^*$,~\IEEEmembership{Member,~IEEE},
        Liansheng~Zhuang,~\IEEEmembership{Member,~IEEE},
        and~Changsheng~Xu,~\IEEEmembership{Fellow,~IEEE}

\thanks{This work was supported by National Science and Technology Major Project(2021ZD0112202), National Natural Science Foundation of China(62376268,U23A20387,62036012,U21B2044,U20B2070), and Beijing Natural Science Foundation (4222039). \emph{(Corresponding author: Changsheng Xu)}}
\thanks{Yukun Zuo and Liansheng Zhuang are with the School of Information Science and Technology, University of Science and Technology of China, Hefei, 230026, China, Email:zykpy@mail.ustc.edu.cn, lszhuang@ustc.edu.cn}
\thanks{Hantao Yao  is with the State Key Laboratory of Multimodal Artificial Intelligence Systems, Institute of Automation, Chinese Academy of Sciences, Beijing, 100190, China, Email:hantao.yao@nlpr.ia.ac.cn}
\thanks{Changsheng Xu is with the State Key Laboratory of Multimodal Artificial Intelligence Systems, Institute of Automation, Chinese Academy of Sciences, Beijing, 100190, China, and also with the School of Artificial Intelligence, University of Chinese Academy of Sciences, Beijing 100049, China, Email: csxu@nlpr.ia.ac.cn}
}

%
%

\markboth{Journal of \LaTeX\ Class Files,~Vol.~14, No.~8, August~2015}%
{Shell \MakeLowercase{\textit{et al.}}: Bare Advanced Demo of IEEEtran.cls for IEEE Computer Society Journals}
%



\IEEEtitleabstractindextext{%
\begin{abstract}
  \justify{
    Audio-visual video recognition (AVVR) aims to integrate audio and visual clues to categorize videos accurately.
    While existing methods train AVVR models using provided datasets and achieve satisfactory results, they struggle to retain historical class knowledge when confronted with new classes in real-world situations.
    Currently, there are no dedicated methods for addressing this problem, so this paper concentrates on exploring Class Incremental Audio-Visual  Video Recognition (CIAVVR).
    For CIAVVR, since both stored  data and learned  model of past classes  contain historical knowledge, the core challenge is how to capture past data knowledge and past model knowledge to prevent catastrophic forgetting.
    As audio-visual data and model inherently contain hierarchical structures, i.e., model embodies low-level and high-level semantic information, and data comprises snippet-level, video-level, and distribution-level spatial information, it is essential to fully exploit the hierarchical data structure  for data knowledge preservation and hierarchical model structure  for model knowledge preservation.
    However, current image class incremental learning methods  do not explicitly  consider these hierarchical structures in model and data.
    Consequently, we introduce Hierarchical Augmentation and Distillation (HAD), which comprises the Hierarchical Augmentation Module (HAM) and Hierarchical Distillation Module (HDM) to efficiently utilize the hierarchical structure of data and models, respectively. 
    Specifically, HAM implements a novel augmentation strategy, segmental feature augmentation, to preserve hierarchical model knowledge.
    Meanwhile, HDM introduces newly designed hierarchical (video-distribution) logical distillation and hierarchical (snippet-video) correlative distillation to capture and maintain the hierarchical intra-sample knowledge of each data  and the hierarchical inter-sample knowledge between data, respectively.
    Evaluations on four benchmarks (AVE, AVK-100, AVK-200, and AVK-400) demonstrate that the proposed HAD effectively captures hierarchical information in both data and models, resulting in better preservation of historical class knowledge and improved performance. 
    Furthermore, we provide a theoretical analysis to support the necessity of the segmental feature augmentation strategy. }
\end{abstract}


}

\maketitle

\IEEEdisplaynontitleabstractindextext

%
\IEEEpeerreviewmaketitle

\ifCLASSOPTIONcompsoc
\IEEEraisesectionheading{\section{Introduction}\label{sec:introduction}}
\else
\section{Introduction}
\label{sec:introduction}
\fi

%
%
%
%

\begin{figure}
  \centering
    \subfigure[Class incremental  image recognition]{
      \includegraphics[width=0.8\linewidth]{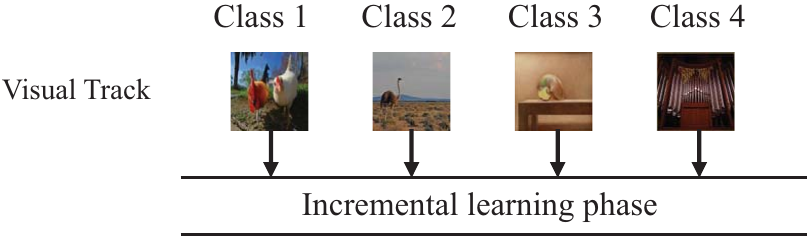}
      }
    \subfigure[Class incremental  audio-visual video recognition]{
      \includegraphics[width=0.8\linewidth]{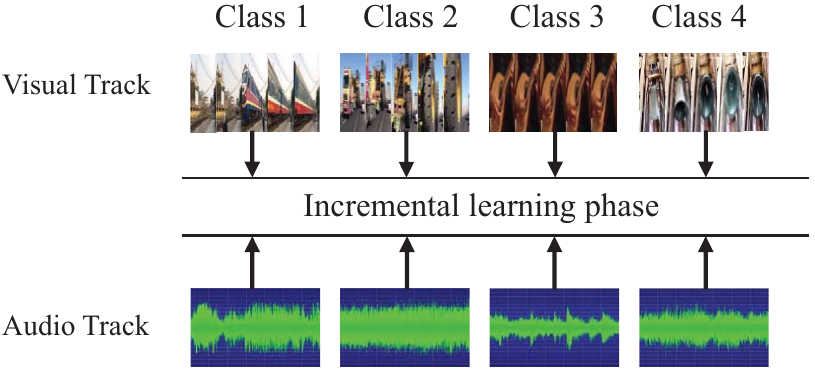}
      }

  \caption{\footnotesize{(a) Most of  class incremental  learning methods focus on image-level knowledge preservation. (b) We focus on class incremental  audio-visual video recognition containing visual information and audio information.}}
  
  \label{task}
  \vspace{-1em}
  \end{figure}

Audio-visual video recognition~\cite{a8,a9,a71,a72,a73} combines audio and visual data for accurate classification and relies on large static datasets of annotated videos for training~\cite{a3,a4}. 
Integrating new class data into these datasets requires significant computational resources but training only on the new class data leads to catastrophic forgetting~\cite{a1, a2}, erasing knowledge about older classes and reducing performance. 
This issue is more challenging in audio-visual recognition due to the richer data involved compared to image recognition shown in Figure~\ref{task}. 
As there has been no specific study addressing catastrophic forgetting in this field, we explore Class Incremental Audio-Visual Video Recognition (CIAVVR) to tackle this issue in audio-visual video recognition.

\begin{figure}
  \begin{center}
  {\includegraphics[width=1\linewidth]{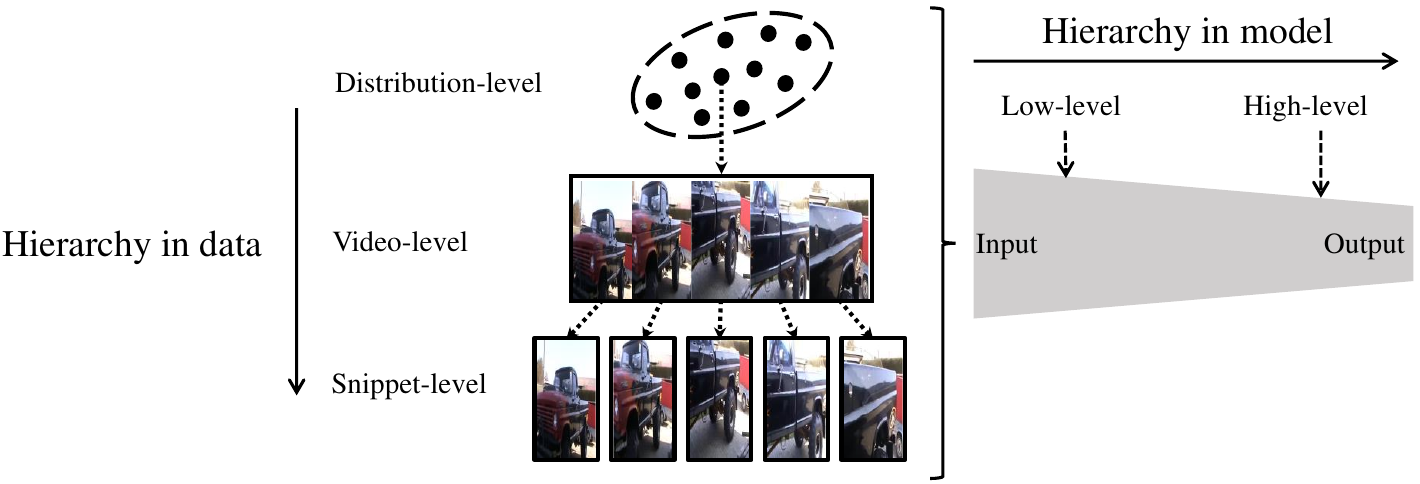}}
  \end{center}
  \caption{The hierarchical structure in  model and video data.  
  For the model, low-level and high-level features embody different semantic information.
  Moreover, the data comprises  distribution-level, video-level, and snippet-level spatial information. 
  }\label{fig_introduction}
  \vspace{-1em}
  \end{figure}

The core idea of CIAVVR is to preserve historical class knowledge from available stored data and  model of past classes to overcome catastrophic forgetting. 
Unlike image tasks, audio-visual tasks involve hierarchical structures within both the model and data shown in Figure~\ref{fig_introduction}.
Specifically,  the low-level and high-level features in the model embody the low-level and high-level semantic knowledge, respectively. 
Furthermore, the video distributions, videos, and snippets in the data comprise distribution-level, video-level, and snippet-level spatial knowledge, respectively.
Thus, fully utilizing the hierarchical structure in data and model is crucial for preserving data knowledge and model knowledge.
However, current image class incremental learning methods do not fully consider these hierarchies, limiting their effectiveness for CIAVVR. Augmentation-based methods~\cite{a11, a6, a14} focus on either low-level or high-level feature augmentation to diversify past data and address class imbalance, but neglect joint hierarchical model learning.
Moreover, merely considering both low-level and high-level feature augmentation leads to an accumulation of augmentation error information. 
Distillation-based methods~\cite{a11, a13, a16} use logical or correlative distillation to capture intra- and inter-sample data knowledge, but fail to characterize knowledge within the hierarchical data structure.

To address catastrophic forgetting in CIAVVR, fully exploiting the hierarchical structure in data and models of past classes is key for preserving both model and data knowledge. 
For  hierarchical model knowledge preservation, we combine low-level and high-level feature augmentation to diversify  exemplar data at various semantic levels, mitigating class imbalance. We also theoretically demonstrate that different levels of feature augmentation specifically influence their respective network layer updates, reducing error accumulation.
For hierarchical data knowledge preservation, we jointly consider hierarchical logical and hierarchical correlative distillation to capture intra-sample and inter-sample knowledge.  
Specifically, we use  logical distillation at both video and distribution levels to grasp hierarchical intra-sample knowledge. Simultaneously, we apply correlative distillation at snippet and distribution levels to understand the hierarchical inter-sample knowledge, focusing on feature similarities among different snippets and videos.

In our study, we introduce the Hierarchical Augmentation and Distillation (HAD) framework for CIAVVR, which includes the Hierarchical Augmentation Module (HAM) and Hierarchical Distillation Module (HDM) for preserving model and data knowledge, respectively. 
For hierarchical model knowledge preservation, HAM employs a novel segmental feature augmentation to enhance stored data generalization through low-level and high-level feature augmentation. 
We prevent interaction between these augmentations in subsequent network layer updates, thus avoiding error information accumulation.
For preserving hierarchical data knowledge, HDM introduces hierarchical logical (video-distribution) and correlative (snippet-video) distillation methods. These maintain intra-sample and inter-sample knowledge respectively. 
Video-distribution logical distillation processes the logical outputs of individual videos and the sampled video from the video distribution. 
Since the video distribution lacks an explicit probability density function, we use the convex hull of the provided data to create a proxy video distribution. 
Additionally, snippet-video correlative distillation focuses on distilling feature correlations between various snippets and videos.

  The contributions of this work are summarized  as follows:
\begin{itemize}
  \item We introduce a novel  Class Incremental  Audio-Visual Video Recognition (CIAVVR) paradigm  for learning from new classes without forgetting old class knowledge using audio-visual information.
  \item We present the Hierarchical Augmentation and Distillation (HAD) framework for CIAVVR, with the Hierarchical Augmentation Module (HAM) and Hierarchical Distillation Module (HDM) for preserving model and data knowledge, respectively.
  \item We develop a new  segmental feature augmentation strategy  in HAM for hierarchical model knowledge, and novel video-distribution logical and snippet-video correlative distillation strategies in HDM for  hierarchical data knowledge. We also provide a theoretical analysis of the segmental feature augmentation necessity.
  \item The evaluations on four benchmarks demonstrate the superiority of the proposed framework, e.g., obtaining Average Incremental Accuracy / Final Incremental Accuracy  of 88.9\%/85.1\% (87.0\%/83.1\%), 90.1\%/86.6\% (89.8\%/86.3\%), 84.6\%/78.0\% (84.3\%/77.6\%) and 78.2\%/69.5\% (77.6\%/69.1\%)  in AVE 3 phases (6 phases), AVK-100 5 phases (10 phases), AVK-200 10 phases (20 phases) and AVK-400 20 phases (40 phases), respectively.
\end{itemize}
  
  \section{Related Work}
  Class incremental audio-visual video recognition is related to three research areas: image-level class incremental learning, video-level class incremental learning, and audio-visual learning.
  In this section, we provide a brief review of these areas.

  \subsection{Image-level Class Incremental  Learning} 
  Image-level class incremental  learning aims to address the problem of catastrophic forgetting for sequential learning in  image-level tasks. 
  Recently, various methods have been proposed to tackle this problem, which can be categorized into non-parametric and parametric methods.
  Moreover, parametric methods can be further classified into three categories: regularization-based methods, modified architecture methods, and exemplar-based methods.

  \subsubsection{Non-parametric Methods}
  \textbf{Non-parametric methods}~\cite{a74,a75,a76,a77,a78} utilize Bayesian inference to retain a distribution for model parameters instead of obtaining the optimal point estimate. 
   VCL~\cite{a74} combines online variational inference and Monte Carlo variational inference to yield variational continual learning.
   BSA~\cite{a75}  jointly considers the adapted structure of deep networks and variational Bayes based regularization together,  and proposes a novel Bayesian framework to  continually learn the deep structure for each task.
   KCL~\cite{a76}  uses kernel ridge regression to learn task-specific classifier to avoid task interference in  the classifier. Moreover, it learns a data-drive informative kernel for each task with variational random features inferred from  the coreset of each task.
   FRCL~\cite{a77} conducts Bayesian inference over the function space instead of model parameters. Specifically, it utilizes inducing point sparse Gaussian process methods to constructs posterior beliefs over the task-specific function, which are memorized  for  functional  regularization to avoid forgetting.
   FRORP~\cite{a78} presents a new functional-regularization method to identify a few memorable past samples which are important to reduce forgetting in Gaussian Process formulation of deep networks. 
   However, the expensive computation of non-parametric methods restricts  their application in complex continual learning scenarios.
  \subsubsection{Parametric Methods}
  \textbf{Regularized-based methods}~\cite{a17,a18,a19,a58,a59} strive to search the important parameters to the original tasks and constrain their changes in the coming tasks. 
  Elastic Weight Consolidation (EWC)~\cite{a17} computes the importance of each parameter via a diagonal approximation of the Fisher Information Matrix, and slows down the learning of each parameter selectively.
  Synaptic Intelligence (SI) ~\cite{a18} breaks the EWC paradigm of calculating the parameter importance in a separate phase, which maintains an online estimation of the changes in each parameter along the entire learning trajectory.
  Different from EWC and SI gaining the parameter by calculating gradients of loss function, Memory Aware Synapses (MAS)~\cite{a19} obtains gradients of $L_2$ norm of the predicted output function to see  how sensitive the predicted output function  to a change for each parameter.
  Riemania Walk (RWalk)~\cite{a58}  calculates the parameter importance by fusing Fisher Information Matrix approximation and online path integral with a theoretically grounded KL-divergence based perspective.
  ELI~\cite{a59} learns an energy manifold for the latent representations to counter the representational shift during Incremental learning. 
  The advantage of the regularized-based method is that they do not need to store the samples of old tasks, \emph{i.e.}, exemplar samples.
  Nonetheless, due to the difficulty of finding essential parameters, the performance of regularized-based methods is not satisfactory.

  \textbf{Modified architectures methods}~\cite{a20,a21,a22,a23,a54,a55,a56, a57} aim to retain knowledge from previous tasks by designing specific network architectures or memorize existing knowledge by adding new network parameters. 
  PNN~\cite{a20} proposes Progressive Networks to retain a pool of pretrained models throughout training, and adds lateral connections between duplicates for each task to extract assistant features for the new task.
  PackNet~\cite{a54} exploits redundancies in a large model, and sequentially packs multiple tasks into the model with minimal drop in performance and minimal storage overhead.
  ACL~\cite{a55} assumes that the representations in each task contain some shared properties and   task-specific properties.
  Therefore, it explicitly disentangles shared and task-specific features with an adversarial loss, and utilizes exemplars to fuse a dynamic architecture.
  DER~\cite{a56} utilizes a dynamically expandable representation in a novel two-stage learning approach for incremental learning. 
  Specifically, it first freezes the learned representation of previous classes, and augments it with additional feature dimensions from a new learnable feature extractor.
  Then, it introduces a channel-level mask-based pruning strategy to dynamically expand the representation.
  FOSTER~\cite{a57} first fits the residuals between the target and the output of the original model  for each new task via expanding new modules dynamically, and then uses a distillation strategy to remove redundant parameters.
  However, these methods suffer from heavy computation for sequential classes, which limits its application in real-world scenarios.

  \textbf{Exemplar-based methods}~\cite{a11, a16, a27,a69} are the current mainstream methods, the core of which is to remember past task knowledge by storing a fraction of historical samples~\cite{a6, a11, a12,a13} or generating the historical samples~\cite{a24,a25,a26}. 
  iCaRL~\cite{a11} utilizes nearest-mean-of-exemplars classifier and herding-based exemplar selection to enforce distillation on new tasks and  historical exemplars for overcoming the catastrophic forgetting.
  LUCIR~\cite{a12} introduces a loose forget constraint and inter-class separation to preserve
  the geometry of past classes and maximize the distances between past and new classes, respectively.
  Furthermore, it utilizes cosine normalization to enforce balanced magnitudes across all classes.
  PODNet~\cite{a60} proposes a spatial-based distillation loss and a representation comprising multiply proxy vectors in each class for knowledge preservation.
  TPCIL~\cite{a61} introduces a Topology-Preserving Class-Incremental Learning framework to maintain the topology in the feature space.
  Specifically, it first constructs the feature topology with the Elastic Hebbian Graph (EHG), and then constrains the changes of neighboring relationships in EHG  via topology-preserving loss (TPL).
  PASS~\cite{a63} memorizes representative prototype for each old class and applies prototype augmentation  to maintain the decision boundary of previous tasks.
  CSCCT~\cite{a62} suggests two distillation-based objectives for class incremental learning, where Cross-space Clustering characterizes the directions of optimization about the feature space structure for  preserving the old classes maximally, and the Controlled Transfer constrains the semantic similarities of incrementally arriving classes and prior classes between  old model and current model. 

  The proposed Hierarchical Augmentation and Distillation (HAD)  belongs to exemplar-based methods. 
  However, current exemplar-based methods  ignore the hierarchical structure in model and data. 
  To solve the above problem, HAD exploits the hierarchical structure in model and data  to implement  model and data knowledge preservation.
  
  \subsection{Video-level Class Incremental  Learning}
  Similar to image-level tasks, video-level tasks are also prone to suffer from catastrophic forgetting.
  TCD~\cite{a80} explores time-channel importance maps in representations and exploits the importance maps of incoming examples for knowledge distillation.  Moreover, it uses a regularization scheme to enforce the features of different time steps to be uncorrelated for further reducing forgetting.
  vCLIMB~\cite{a80} observes two unique challenges in video-level class incremental  learning, \emph{i.e.}, the selection of instances in memory are sampled in the frame  level and the effectiveness of frame sampling is affected by untrimmed videos. It proposes a temporal consistency regularization to overcome these two challenges.
  FrameMaker~\cite{a82} presents a memory effective video-level class incremental  learning approach consisting of Frame Condensing and Instance-Specific Prompt. Specifically, Frame Condensing preserves one condensed frame rather than individual video, and Instance-Specific Prompt compensates the lost spatio-temporal details of the condensed frame.
  However, the above methods cannot handle the multimedia information and hierarchical structure  existed in class incremental  audio-visual video recognition.

  \subsection{Audio-visual Learning}
  Video contains audio and visual modal information naturally, which have strong semantic correlation.
  However, many methods~\cite{a64,a65,a66,a67} only focus the exploration and exploitation of visual information  while ignoring the significant audio clues. 
  Moreover, the temporal alignment between audio and visual data benefits to learn the video representation.
  Therefore, how to utilize and fuse audio and visual information in videos effectively is crucial for the application of video representation learning.
  Audio-visual learning has been widely studied and applied in many areas, such as audio-visual representation learning~\cite{a35,a36,a37,a38,a53}, audio-visual event localization~\cite{a28,a29,a30,a31,a51,a52}, audio-visual source separate~\cite{a32,a33,a34},  audio-visual video captioning~\cite{a42,a43,a44}, and audio-visual action recognition~\cite{a39,a40,a41}.
  However, most of these methods assume that the data of all tasks are static and available, ignoring that the data may be given via sequence flows in real-world scenarios.
  Different from the above methods, we study a fundamental task in audio-visual learning: audio-visual video recognition, and explore the problem of learning the knowledge with sequential data named class incremental  audio-visual video recognition.
  CIAVVR aims to learn to recognize new audio-visual video classes without forgetting the knowledge of old audio-visual video classes.
  
  \begin{figure*}
      \begin{center}
      {\includegraphics[width=0.9\linewidth]{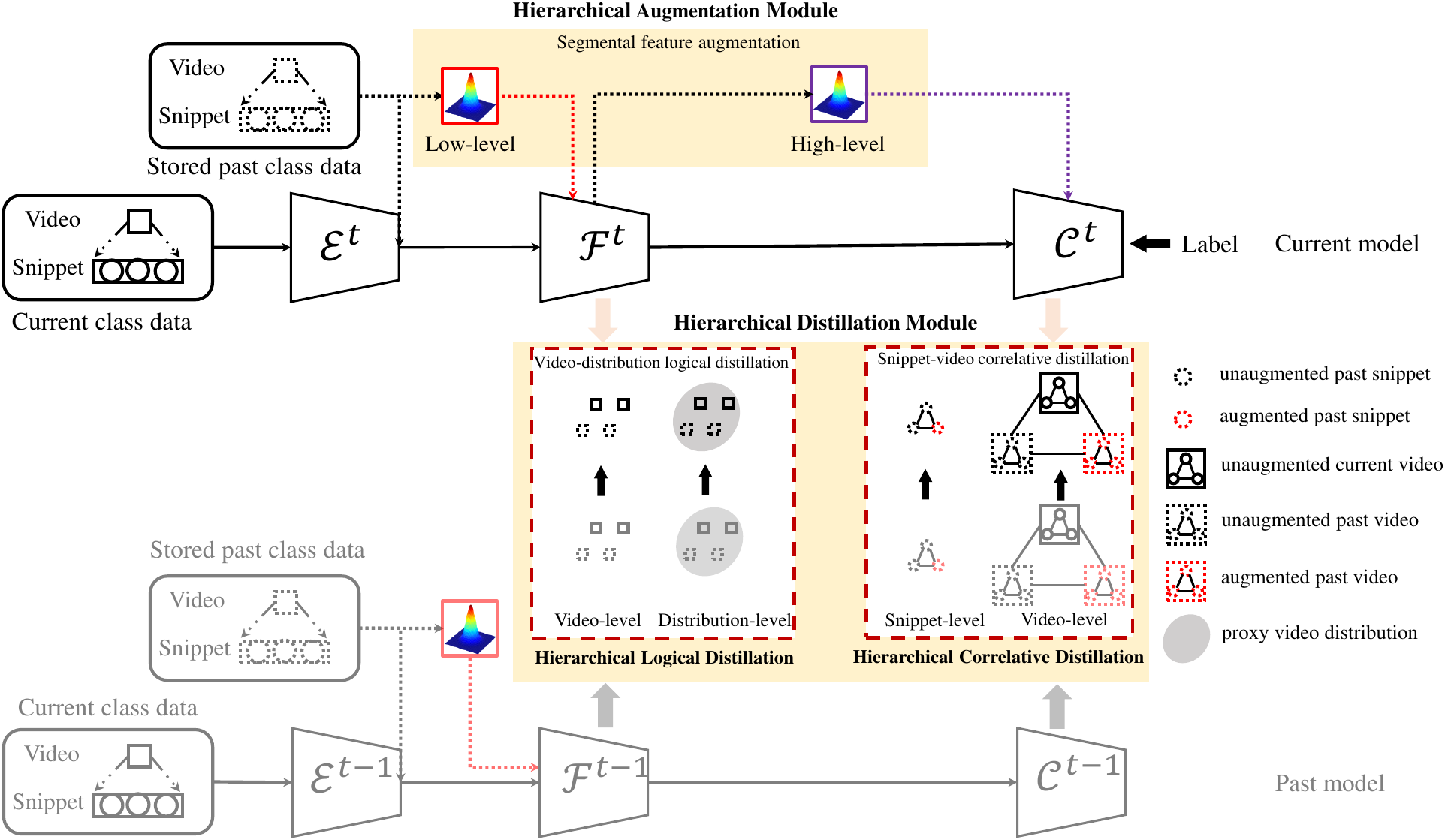}}
      \end{center}
      \caption{The proposed Hierarchical Augmentation and Distillation (HAD) framework consists of Hierarchical Augmentation Module (HAM) and Hierarchical Distillation Module (HDM). HAM utilizes \emph{segmental feature augmentation} to conduct the low-level and high-level feature augmentations for enhancing data knowledge preservation.
      Moreover, HDM consisting of  hierarchical logical distillation (HLD) and hierarchical correlative distillation (HCD) employs \emph{video-distribution logical distillation} and \emph{snippet-video correlative distillation} for model knowledge preservation. 
      }
      \label{framework}
      \end{figure*}
  
  \section{Methods}
  \subsection{Problem Definition.} 
  Class Incremental  Audio-Visual Video Recognition (CIAVVR) aims to learn new audio-visual video classes without forgetting the knowledge of old classes.
  Formally, given a sequence of $S$ tasks $\{\mathcal{T}_1,\mathcal{T}_2,\cdots, \mathcal{T}_S\}$, and each task $\mathcal{T}_t$ has a task-specific class set $\mathcal{Y}_t$, where different tasks have disjoint class sets, \emph{i.e.}, $\mathcal{Y}_i \cap \mathcal{Y}_j = \emptyset$, if $i \neq j$.
  To retain knowledge from previous tasks, at the step $t$, a small amount of exemplar data from previous tasks $\mathcal{T}_{1:(t-1)}$ is stored in a memory bank $\mathcal{M}_{t-1}$ with a limited size $M$. 
  For the $t$-th task, CIAVVR constructs a robust audio-visual model using the dataset $\mathcal{T}_t = \{(x,y)|(x,y) \in \mathcal{D}_t\}$ and the exemplar data in memory bank $\mathcal{M}_{t-1}$, where $x \in \mathcal{X}$ denotes a video  sampled from the  video  space $\mathcal{X}$, $y \in \mathcal{Y}_t$ represents its video label belonging to task-specific class set $\mathcal{Y}_t$, and $\mathcal{D}_t$ signifies the joint distribution of video $x$ and label $y$.
  The inferred audio-visual model must accurately classify the testing datasets $\{\mathcal{T}_1,\mathcal{T}_2,\cdots, \mathcal{T}_t\}$ among all previous $t-1$ tasks and the current  $t$-th task at the incremental step $t$.

  Given a video $x$, we divide it into $K$ disjoint audio and visual snippet pairs, \emph{e.g.,} $x=\{A_i, V_i\}_{i=1}^K$, where $A_i$ and $V_i$ represent the audio and visual data of the $i$-th video snippet, respectively.
  The audio-visual model aims to classify the video $x$ by considering all audio snippets $\{A_i\}_{i=1}^K$ and visual snippets $\{V_i\}_{i=1}^K$.
  The audio-visual model $\Phi$ comprises three components: the audio-visual embedding module $\mathcal{E}$, the audio-visual fusion module $\mathcal{F}$, and the classifier module $\mathcal{C}$. 
  The audio-visual embedding module $\mathcal{E}$ employs  pre-trained and frozen audio and visual models to extract the low-level modal features $\mathbb{F} = \{f_i^a, f_i^v\}_{i=1}^K$, which consist of audio snippet-level feature $f_i^a$ and visual snippet-level feature $f_i^v$.
  Therefore, the primary emphasis of  CIAVVR is on the learning and catastrophic forgetting of the audio-visual fusion module and classifier module, rather than on the catastrophic forgetting of the frozen pre-trained audio-visual embedding module.
  Since storing the videos of past classes demands significant memory, low-level modal features of exemplar videos are retained for knowledge preservation.
  Specifically, for incremental learning at the $t$-th task, the low-level modal features $\mathbb{F}^{m}$ of the exemplar data $x^m$ from the previous $t-1$  tasks are stored in the memory bank $\mathcal{M}_{t-1}$ for knowledge preservation.
  Moreover, we denote the set of low-level modal features $\mathbb{F}^c$ of  current data $x^c$  in $\mathcal{T}_t$ as $\mathcal{T}'_t$.
  The audio-visual fusion module $\mathcal{F}$ utilizes the hybrid attention network~\cite{a29} to perform multi-modal fusion between  $\{f_i^a\}_{i=1}^K$ and $\{f_i^v\}_{i=1}^K$, yielding fused audio features $\{h_i^a\}_{i=1}^K$ and visual features $\{h_i^v\}_{i=1}^K$.
  The video feature $H$ is obtained by fusing the  visual and audio features  using average pooling,
  \begin{align}
      H = H^a + H^v = \frac{1}{K}\sum_{i=1}^{K} h_i^a +\frac{1}{K}\sum_{i=1}^{K}h_i^v,
  \label{eq:video_feature}
  \end{align}
  where $H^a = \frac{1}{K}\sum_{i}^{K} h_i^a$ and $H^v = \frac{1}{K}\sum_{i}^{K} h_i^v$ represent the video-level audio feature and visual feature, respectively.

  With the obtained video-level feature $H$, the classifier $\mathcal{C}$ predicts its categories.
  The goal of audio-visual model is to accurately classify the videos of all previous $t-1$ tasks with the data of $t$-task,
  \begin{align}
    \mathcal{L} = \mathbb{E}_{(x,y) \in \mathcal{T}_t}[l_{ce}(y, \mathcal{C}(\mathcal{F}(\mathcal{E}(x))))],
  \label{eq:base}
  \end{align}
  where $l_{ce}$ denotes the cross-entropy loss.
  
  The critical challenge of CIAVVR is how to preserve the knowledge of the old tasks while training the new task.
  We thus propose a novel Hierarchical Augmentation and Distillation (HAD) framework for CIAVVR, consisting of Hierarchical Augmentation Module (HAM) and Hierarchical Distillation Module (HDM) to preserve model knowledge and data knowledge via the hierarchical structure of model and video data, respectively, as shown in Figure~\ref{framework}.

  \subsection{Hierarchical Augmentation Module}
  Hierarchical Augmentation Module explores the hierarchical structure in model for model knowledge preservation. 
  We  propose a novel \emph{segmental feature augmentation} strategy to concurrently augment  old exemplars $\mathcal{M}_{t-1}$ from both low-level and high-level perspectives,  preserving the historical model knowledge and enhancing the generalization of the model.
  Additionally, to alleviate error information accumulation caused by augmentation, we make different levels of feature augmentation  update the parameters of different modules.

  Given the low-level modal features $\mathbb{F}^{m} = \{f_i^{m,a}, f_i^{m,v}\}_{i=1}^K$ of the historical samples in the memory bank $\mathcal{M}_{t-1}$, we perform Gaussian augmentation $z \sim \mathcal{N}(\mathbf{0}, \mathbf{1})$ to generate the augmented modal features $\bar{\mathbb{F}}^{m} = \{\bar{f}_i^{m,a}, \bar{f}_i^{m,v}\}_{i=1}^K$:
  \begin{align}
    \bar{f}^{m,a}_i = f_i^{m,a} + \lambda * z, \quad
    \bar{f}^{m,v}_i = f_i^{m,v} + \lambda * z,
  \label{eq:1}
  \end{align}
  where $\lambda$  represents the intensity of Gaussian  augmentation.
  For the low-level augmented modal feature $\bar{\mathbb{F}}^{m}$,
  it is  only applied to update the audio-visual fusion module $\mathcal{F}$ by fixing the classifier $\mathcal{C}$:
  \begin{align}
    \mathcal{L}_{lsm}(\varphi) =  \mathbb{E}_{(\mathbb{F}^{m},y^m) \in \mathcal{M}_{t-1}}[l_{ce}(y^m, \mathcal{C}(\mathcal{F}(\bar{\mathbb{F}}^{m}))],
  \label{eq:1}
  \end{align}
  where $\varphi$ is the parameter of the audio-visual fusion module $\mathcal{F}$, and $\mathcal{L}_{lsm}(\varphi)$ denotes only updating the parameter $\varphi$.

  To mitigate the impact of class imbalance for classifier $\mathcal{C}$, the high-level video feature augmentation is employed to update the classifier.
  Given $\mathbb{F}^{m}$ of historical samples, we firstly apply the audio-visual fusion module to generate the corresponding video-level features $H^m$ with Eq.~\eqref{eq:video_feature}. 
  We then perform Gaussian augmentation $z \sim \mathcal{N}(\mathbf{0}, \mathbf{1})$ on the video-level features $H^m$ for high-level video feature augmentation:
  \begin{align}
    \bar{H}^m = H^m + \lambda * z.
  \label{eq:1}
  \end{align}
  Subsequently, the augmented  feature $\bar{{H}}$ is used to optimize the classifier $\mathcal{C}$:
  \begin{align}
    \mathcal{L}_{hsm}(\psi) =  \mathbb{E}_{(\mathbb{F}^m,y^m) \in \mathcal{M}_{t-1}}[l_{ce}(y^m, \mathcal{C}(\bar{H}^m)],
  \label{eq:1}
  \end{align}
  where $\psi$ denotes the parameter of the classifier $\mathcal{C}$.
  
  Finally, the total loss of HAM is:
  \begin{align}
    \mathcal{L}_{HAM} = \mathcal{L}_{lsm}(\varphi) +\mathcal{L}_{hsm}(\psi).
  \label{eq:1}
  \end{align}

  \subsection{Hierarchical Distillation Module}
  Apart from considering the hierarchical model structure for model knowledge preservation, the hierarchical  structure within data also can be explored to preserve data knowledge.
  Leveraging the hierarchical structure present in data, we introduce a Hierarchical Distillation Module (HDM) to maintain  historical data knowledge,  reducing  catastrophic forgetting.
  HDM consists of Hierarchical Logical Distillation (HLD) and Hierarchical Correlative Distillation (HCD).
  HLD is employed to distill the logical probability of each given video and the sampled video from the video distribution.
  HCD is responsible for distilling feature similarities between different snippets in each video and different videos in the video space.

  \subsubsection{Hierarchical Logical Distillation}
  Given the historical memory $\mathcal{M}_{t-1}$ and the data $\mathcal{T}_t$ of the current task, we perform the \emph{video-distribution logical distillation} based on the predicted logical probability between the historical model $\Phi^{t-1}$ and current model $\Phi^{t}$.
  
  Since the historical memory only stores the low-level modal description of the historical samples, the model $\mathcal{CF}$, consisting of the audio-visual fusion module $\mathcal{F}$ and the classifier module $\mathcal{C}$, is employed to distill the knowledge of low-level modal description.
  Specifically, by fixing the historical model $\mathcal{CF}^{t-1}$, we constrain the current model $\mathcal{CF}^{t}$ to produce the consistent logical probability with the past model based on the video-level logical distillation $\mathcal{L}_{sl}$: 
  \begin{align}
    \mathcal{L}_{sl} &=  \mathbb{E}_{\mathbb{F}^m\in \mathcal{M}_{t-1}}[l_{kl}(\mathcal{CF}^{t-1}(\mathbb{F}^m), \mathcal{CF}^{t}(\mathbb{F}^m))]  \notag\\ &+ \mathbb{E}_{\mathbb{F}^c \in \mathcal{T}'_t}[l_{kl}(\mathcal{CF}^{t-1}(\mathbb{F}^c), \mathcal{CF}^{t}(\mathbb{F}^c)))],
  \label{eq:1}
  \end{align}
  where $l_{kl}$ represents the Kullback-Leibler divergence. 
  
  However, the video-level logical distillation only focuses the individual  knowledge of each video, ignoring the underlying individual knowledge of sampled video from   video distribution.  
  Ideally, we want to distill the knowledge of any video sampled from the old task distribution $\mathcal{D}_{1:t-1} = {\mathcal{D}_1 \cup \mathcal{D}_2 \cup \cdots \cup \mathcal{D}_{t-1}}$ and current task distribution $\mathcal{D}_t$:
  \begin{align}
    \mathcal{L}_{dl} &=  \mathbb{E}_{x\backsim \mathcal{D}_{1:t-1}}[l_{kl}(\Phi^{t-1}(x), \Phi^{t}(x))]  \notag\\&+ \mathbb{E}_{x\backsim \mathcal{D}_{t}}[l_{kl}(\Phi^{t-1}(x), \Phi^{t}(x))].
  \label{eq:1}
  \end{align}
  
  Therefore, we utilize all the given  low-level modal features in the memory bank $\mathcal{M}_{t-1}$ and current task $\mathcal{T}'_t$ to obtain proxy distribution of $\mathcal{D}_{1:t-1}$ and $\mathcal{D}_{t}$.
  Specifically, the proxy distributions of $\mathcal{D}_{1:t-1}$ and $\mathcal{D}_{t}$ are constructed by the \emph{convex hull} $\mathcal{H}(\mathcal{M}_{t-1})$ and $\mathcal{H}(\mathcal{T}'_t)$, using all convex combinations of the low-level modal features in the set $\mathcal{M}_{t-1}$ and set $\mathcal{T}'_t$, respectively:
  \begin{align}
    \mathcal{H}(\mathcal{M}_{t-1}) = \{\sum_{i=1}^{|\mathcal{M}_{t-1}|} \alpha_i \mathbb{F}^m_i&|\mathbb{F}^m_i \in \mathcal{M}_{t-1}, \notag\\&\sum_{i=1}^{|\mathcal{M}_{t-1}|} \alpha_i = 1,  \alpha_i \in [0,1]\},
    \label{eq:1}
  \end{align}
  \begin{align}
    \mathcal{H}(\mathcal{T}'_t) = \{\sum_{i=1}^{|\mathcal{T}'_t|} \alpha_i \mathbb{F}^c_i|\mathbb{F}^c_i \in  \mathcal{T}'_t, \sum_{i=1}^{|\mathcal{T}'_t|} \alpha_i = 1, \alpha_i \in [0,1]\},
  \label{eq:1}
  \end{align}
  where $|\mathcal{M}_{t-1}|$ and $|\mathcal{T}'_t|$ represent the number of low-level modal features $\mathbb{F}^m$ and $\mathbb{F}^c$  in the set $\mathcal{M}_{t-1}$ and $\mathcal{T}'_t$, respectively.
  $\alpha_i$ depicts the weight of $i$-th low-level modal feature $\mathbb{F}^m_i$ or $\mathbb{F}^c_i$. The weights are first sampled from Gaussian distribution $\mathcal{N}(0,1)$, and then are normalized to [0,1] for convex combination.  
  By adjusting the value of convex combination weight  $\alpha_i$, we simulate the low-level modal features of different videos in the data distribution $\mathcal{D}_{1:t-1}$ and $\mathcal{D}_{t}$.  
  Note that due to GPU memory limitations, we use the low-level modal features of batch samples in current / past task data to conduct convex combination for simulating the low-level modal features of all videos sampled from the data distribution $\mathcal{D}_{t}$ / $\mathcal{D}_{1:t-1}$ in each epoch.

  The data sampled from $\mathcal{H}(\mathcal{M}_{t-1})$ and $\mathcal{H}(\mathcal{T}_t)$ are used for logical distillation with Eq.~\eqref{eq:ld}.
  \begin{align}
    \mathcal{L}_{dl} &=  \mathbb{E}_{\mathbb{F}^m\in \mathcal{H}(\mathcal{M}_{t-1})}[l_{kl}(\mathcal{CF}^{t-1}(\mathbb{F}^m), \mathcal{CF}^{t}(\mathbb{F}^m))] \notag \\&+ \mathbb{E}_{\mathbb{F}^c \in \mathcal{H}(\mathcal{T}'_t)}[l_{kl}(\mathcal{CF}^{t-1}(\mathbb{F}^c), \mathcal{CF}^{t}(\mathbb{F}^c)))], 
  \label{eq:ld}
  \end{align}
  
  The total loss of Hierarchical Logical Distillation  is:
  \begin{align}
    \mathcal{L}_{HLD} =  \mathcal{L}_{sl} + \mathcal{L}_{dl}.
  \label{eq:1}
  \end{align}

  \subsubsection{Hierarchical Correlative Distillation}
  Hierarchical Logical Distillation (HLD) accounts for preserving historical individual knowledge by considering hierarchical intra-sample knowledge. However, it neglects the preservation of historical correlation knowledge derived from inter-sample knowledge, which can also contribute to data knowledge preservation. We propose Hierarchical Correlative Distillation to address this by distilling feature similarities between different snippets in each video and different videos in the video space, \emph{i.e.,} snippet-level correlative knowledge and video-level correlative knowledge. Furthermore, since each video contains audio and visual information, we consider hierarchical correlative distillation for each modality. To simplify the description, we omit the modal identifier in the following discussion.

  For video-level  correlative distillation, we utilize the similarity of the video-level feature between each augmented sample $\bar{\mathbb{F}}^{m}$ in $\mathcal{M}_{t-1}$  and the unaugmented samples $\mathbb{F}$ in $\mathcal{M}_{t-1} \cup \mathcal{T}'_t$ to represent the video-level  feature correlation:
  \begin{align}
    s_{ij} =  \frac{\exp(\bar{H}^{m}_i \cdot H_j)}{\sum_{\mathbb{F}_k \in \mathcal{M}_{t-1} \cup \mathcal{T}'_t}\exp(\bar{H}^{m}_i \cdot H_k)},
  \label{eq:1}
  \end{align}
  where $s_{ij}$ denotes the similarity of the video-level  features between the  $i$-th augmented low-level modal feature $\bar{\mathbb{F}}^{m}_i$ in  $\mathcal{M}_{t-1}$ and the $j$-th unaugmented low-level modal feature $\mathbb{F}_j$ in $\mathcal{M}_{t-1} \cup \mathcal{T}'_t$,
  $\bar{H}^{m}_i$ represents the video-level  feature of the $i$-th augmented low-level modal feature $\bar{\mathbb{F}}^{m}_i$ in  $\mathcal{M}_{t-1}$, and $H_j$ depicts the  video-level  feature of the $j$-th unaugmented low-level modal feature $\mathbb{F}_j$ in $\mathcal{M}_{t-1} \cup \mathcal{T}'_t$.
  Furthermore, the video-level  correlative distillation $\mathcal{L}_{ss}$ is conducted using the video similarity matrices between the historical model $\Phi^{t-1}$ and current model $\Phi^{t}$:
  \begin{gather}
    \mathcal{L}_{ss} =  l_{kl}(S^{t-1}, S^{t}), 
  \label{eq:1}
  \end{gather}
  where $S^{t-1} \in \mathbb{R}^{|\mathcal{M}_{t-1}| \times |\mathcal{M}_{t-1} \cup \mathcal{T}_t|}$ and $S^{t} \in \mathbb{R}^{|\mathcal{M}_{t-1}| \times |\mathcal{M}_{t-1} \cup \mathcal{T}_t|}$ represent the video similarity matrices consisting of $s^{t-1}_{ij}$ and $s^t_{ij}$ in the fixed historical model $\Phi^{t-1}$ and current model $\Phi^{t}$,  respectively. 
  
  For snippet-level  correlative distillation, we compute the similarity between the augmented snippet's  fused feature $\{\bar{h}^m_i\}_{i=1}^K$  and the unaugmented snippet's fused features $\{h^m_i\}_{i=1}^K$ in $\mathcal{M}_{t-1}$  to represent the snippet-level feature correlation:
  \begin{align}
    q_{ij} =  \frac{\exp(\bar{h}^m_i\cdot h^m_j)}{\sum_{k \in [1:K]}\exp (\bar{h}^m_i\cdot h^m_k)},
  \label{eq:1}
  \end{align}
  where $q_{ij}$ denotes the similarity between the $i$-th augmented snippet's  fused feature $\bar{h}^m_i$ and the $j$-th unaugmented snippet's fused feature $h^m_j$.
  
  Following that, we conduct the snippet-level  correlative distillation $\mathcal{L}_{ns}$ via snippet similarity matrices between the fixed historical model $\Phi^{t-1}$ and current model $\Phi^{t}$ for all samples in $\mathcal{M}_{t-1}$:
  \begin{gather}
    \mathcal{L}_{ns} =  \mathbb{E}_{\mathbb{F}^m \in \mathcal{M}_{t-1}} l_{kl}(Q^{t-1}(\mathbb{F}^m) , Q^{t}(\mathbb{F}^m)), 
  \label{eq:1}
  \end{gather}
  where $Q^{t-1}(\mathbb{F}^m) \in \mathbb{R}^{K \times K}$ and $Q^{t}(\mathbb{F}^m) \in \mathbb{R}^{K \times K}$ are the snippet similarity matrices in $\mathbb{F}^m$  consisting of $q^{t-1}_{ij}$ and $q^{t}_{ij}$ in historical model $\Phi^{t-1}$ and current model $\Phi^{t}$,  respectively. 
  
  Combining the hierarchical correlative distillation in audio modal and visual modal jointly,
  the total loss of Hierarchical Correlative Distillation  is:
  \begin{align}
    \mathcal{L}_{HCD} =  \mathcal{L}^a_{ss} + \mathcal{L}^a_{ns} + \mathcal{L}^v_{ss} + \mathcal{L}^v_{ns}.
  \label{eq:1}
  \end{align}

  \subsection{Overall Objective}
  The total objective function of the Hierarchical Augmentation and Distillation framework combines the Hierarchical Augmentation Module and Hierarchical Distillation Module:
  \begin{align}
    \mathcal{L}_{ALL} =  \mathcal{L}_{cls} + \beta \mathcal{L}_{HAM} + \gamma \mathcal{L}_{HLD} + \eta \mathcal{L}_{HCD} ,
  \label{eq:1}
  \end{align}
  where $\beta$, $\gamma$, and $\eta$ are the trade-off parameters, $\mathcal{L}_{cls} = \mathbb{E}_{(x,y) \in \mathcal{T}_t}[l_{ce}(y, \mathcal{C}^{t}(\mathcal{F}^{t}(\mathcal{E}^{t}(x))))]$ represents the supervised loss about the current task data in $\mathcal{T}_t$, $\mathcal{L}_{HAM}$ denotes the loss of hierarchical augmentation module, $\mathcal{L}_{HLD}$ and $\mathcal{L}_{HCD}$ indicate the losses of hierarchical logical distillation and hierarchical correlative distillation in hierarchical distillation module, respectively.

  \section{Theoretical Analysis for Hierarchical Augmentation}
  Hierarchical Augmentation Module (HAM) simultaneously considers the low-level modal augmentation and high-level video augmentation to preserve the data knowledge.
  To mitigate the error information caused by augmentation, we assume that different levels of augmentation strategies update different modules, \emph{i.e.}, low-level modal augmentation and high-level video feature augmentation update the parameters of the audio-visual fusion module $\mathcal{F}$ and classifier module $\mathcal{C}$, respectively.
  We then provide a theoretical analysis of HAM to demonstrate its effectiveness in preserving knowledge.

  \subsection{Effect of Augmentation}
  In this section, we prove that using low-level modal augmentation and high-level video augmentation is beneficial for learning the audio-visual fusion module $\mathcal{F}$ and classifier module $\mathcal{C}$.

  Given network weights $w$, dataset $\mathcal{T}$, prior distribution over network weights $p(w)$,  likelihood function $p(\mathcal{T}|w)$, and normalizing constant $p(\mathcal{T})$, we have:
  \begin{align}
  p(w|\mathcal{T}) &= \frac{p(\mathcal{T}|w)p(w)}{p(\mathcal{T})} \notag \\
          &= \frac{p(\mathcal{T}|w)p(w)}{\int p(\mathcal{T}|w)p(w) \mathrm{d}w}.
  \end{align}

 Assuming $w^*$ is the optimal parameter for the given dataset $\mathcal{T}$, we have:
  \begin{align}
      \log p(w^*|\mathcal{T}) = \log p(\mathcal{T}|w^*) + \log p(w^*) - \log p(\mathcal{T}).
      \label{eq:2}
  \end{align}
  
  For $p(\mathcal{T})$, we  have $p(\mathcal{T}) = \int p(\mathcal{T}|w)p(w) \mathrm{d}w < \int p(\mathcal{T}|w^*)p(w) \mathrm{d}w = p(\mathcal{T}|w^*)$. 
  Assuming the augmented dataset $\mathcal{T}'$ is close to the data distribution of $\mathcal{T}$, we then  have $p(\mathcal{T}') < p(\mathcal{T}'|w^*)$
  
  After integrating the given dataset $\mathcal{T}$ and the augmented dataset $\mathcal{T}'$, $\hat{\mathcal{T}} = \mathcal{T} \bigcup \mathcal{T}'$, we have:
  \begin{align}
    &\log p(w^*|\hat{\mathcal{T}}) \notag \\&= \log p(\hat{\mathcal{T}}|w^*) + \log p(w^*) - \log p(\hat{\mathcal{T}})  \\
                          &= \log p(\mathcal{T}, \mathcal{T}'|w^*) + \log p(w^*) - \log p(\mathcal{T},\mathcal{T}')  \\
                          &= \log p(\mathcal{T}|w^*) + \log p(\mathcal{T}'|w^*) + \log p(w^*) \notag \\&- \log p(\mathcal{T}) - \log p(\mathcal{T}')  \\
                          &= \log p(\mathcal{T}|w^*)  + \log p(w^*) - \log p(\mathcal{T}) \notag \\&+ \log p(\mathcal{T}'|w^*) - \log p(\mathcal{T}')  \\
                          &>  \log p(\mathcal{T}|w^*) + \log p(w^*) - \log p(\mathcal{T}) \label{eq:1}  \\
                          &= \log p(w^*|\mathcal{T}).
  \end{align}
  Eq.~\eqref{eq:1} holds since $p(\mathcal{T}'|w^*) > p(\mathcal{T}')$.
  As $p(w^*|\hat{\mathcal{T}}) > p(w^*|\mathcal{T})$, we can obtain a more reasonable Maximum a-posteriori estimation (MAP) for optimal parameter $w^*$ when utilizing the augmented dataset $\mathcal{T}'$  for training, which demonstrates the effectiveness of augmentation for network optimization.

  \subsection{Effect  of  Hierarchical Augmentation}
  In this section,  we demonstrate that making low-level modal feature augmentation and high-level video feature  augmentation separately update  $\mathcal{F}$ and  $\mathcal{C}$  is more effective than using low-level modal augmentation to update $\mathcal{F}$ and  $\mathcal{C}$.
  
  \textbf{Definition.} Given two metric spaces $(X, d_X)$ and $(Y, d_Y)$, where  $d_X$ denotes the metric on the set $X$ and $d_Y$ is the metric on set $Y$, a function $f$ : $X \to Y$ is called \textbf{Lipschitz continuous}\cite{a48} if there exists a real constant $K \ge 0$ such that, for all $x_1$ and $x_2$ in $X$:
  \begin{equation}
      \small
      \begin{aligned}
          d_Y(f(x_1), f(x_2)) \le K d_X(x_1, x_2).
          \label{eq:lipschitz}
      \end{aligned}
      \end{equation}
  When $K = 1$, Eq.~\eqref{eq:lipschitz} is referred as 1-Lipschitz continuous.
  However, Lipschitz continuous is a too strict constraint for deep neural network, \emph{i.e.}, if the function $f_{w^*}$ represents deep neural network,  $d_Y(f_w^*(x_1), f_w^*(x_2)) < K d_X(x_1, x_2)$ does not always holds, \emph{e.g.}, LCSA\cite{a49} proves that standard dot-product self-attention is not Lipschitz continuous for unbounded input domain.
  Therefore, we  assume that the audio-visual fusion module $\mathcal{F}$  does not satisfy the 1-Lipschitz continuous.

  As the audio-visual embedding module $\mathcal{E}$ is frozen in the audio-visual model $\Phi$, we only concentrate on the audio-visual fusion module $\mathcal{F}$ and classifier module $\mathcal{C}$.
  We denote the low-level modal feature as $\mathcal{T}$, and $\mathcal{F}(\mathcal{T})$ represents the high-level video feature.
  We split the neural network $f_{w^*}$ into two part: audio-visual fusion module $\mathcal{F}$ and classifier module $\mathcal{C}$ 
  , whose parameters are $w^*_\mathcal{F}$ and $w^*_\mathcal{C}$, respectively.
  With the augmented dataset $\mathcal{T}'$, we have:
  \begin{align}
  \log p(w^*|\mathcal{T}') &= \log p(w^*_\mathcal{F}, w^*_\mathcal{C}|\mathcal{T}') \notag \\
                      &= \log p(w^*_\mathcal{F}|\mathcal{T}') + \log p(w^*_\mathcal{C}|\mathcal{T}') \notag\\
                      &= \log p(w^*_\mathcal{F}|\mathcal{T}') + \log p(w^*_\mathcal{C}|\mathcal{F}(\mathcal{T}')).
  \label{eq:5}
  \end{align}
  
  Since $\mathcal{T}'$ is the augmentation of the dataset $\mathcal{T}$, it has a similar distribution with $\mathcal{T}$.
  Because audio-visual fusion module $\mathcal{F}$ is not 1-Lipschitz continuous, the high-level video feature $\mathcal{F}(\mathcal{T}')$ of $\mathcal{T}'$ is easy to be away from the  distribution of $\mathcal{F}(\mathcal{\mathcal{T}})$, which can be formulated as follows:
  \begin{align}
          &\log p(w^*_\mathcal{C}|\mathcal{F}(\mathcal{T}')) \notag\\&= \log p(\mathcal{F}(\mathcal{T}')|w^*_\mathcal{C}) + \log p(w^*_\mathcal{C}) - \log p(\mathcal{F}(\mathcal{T}')) \\
          &\approx \log p(\mathcal{F}(\mathcal{T}')|w^*_\mathcal{C}) + \log p(w^*_\mathcal{C}) - \log p({\mathcal{F}(\mathcal{T})}') \label{eq:2}\\
          &< \log p({\mathcal{F}(\mathcal{T})}'|w^*_\mathcal{C}) + \log p(w^*_\mathcal{C}) - \log p({\mathcal{F}(\mathcal{T})}') \\
          &= \log p(w^*_\mathcal{C}|{\mathcal{F}(\mathcal{T})}').
          \label{eq:6}
      \end{align}
  
  Eq.~\eqref{eq:2} holds since for normalizing constant we have $p(\mathcal{F}(\mathcal{T}')) \approx p({\mathcal{F}(\mathcal{T})}')$.
  Therefore,   ${\mathcal{F}(\mathcal{T})}'$  obtains a more reasonable Maximum a-posteriori estimation (MAP) for optimizing the classifier parameter $w^*_\mathcal{C}$ than $\mathcal{F}(\mathcal{T}')$.
  
  After combining Eq.~\eqref{eq:5} and Eq.~\eqref{eq:6}, we conclude:
  \begin{equation}
      \small
      \begin{aligned}
          \log p(w^*|\mathcal{T}') &= \log p(w^*_\mathcal{F}|\mathcal{T}') + \log p(w^*_\mathcal{C}|\mathcal{F}(\mathcal{T}')) \\
          &< \log p(w^*_\mathcal{F}|\mathcal{T}') + \log p(w^*_\mathcal{C}|{\mathcal{F}(\mathcal{T})}')
          \label{eq:7}
      \end{aligned}
      \end{equation}
  Therefore, jointly augmenting  $\mathcal{T}$ and ${\mathcal{F}(\mathcal{T})}$ is more effective than merely augmenting  $\mathcal{T}$.
  Moreover,  the parameters of the audio-visual fusion module $w^*_\mathcal{F}$ and the classifier module $w^*_\mathcal{C}$ should be updated by the augmented samples $\mathcal{T}'$ and augmented features  ${\mathcal{F}(\mathcal{T})}'$, respectively.

\begin{table*}
    \centering
    \scriptsize
    \setlength{\tabcolsep}{8.0pt}
    \footnotesize
    \renewcommand{\arraystretch}{1.2}
    \caption{Dataset statistics and evaluation protocol for the AVE, AVK-100, AVK-200, and AVK-400 datasets.}
    \label{data_statistics}
    
    {
      \begin{tabular}{c c c c c c c c c}
        \shline
        Dataset & \#Total class & \#Train sample & \#Valid sample &\#Test sample & \#Task & \#Inital class& \#Incremental class & \#Memory size\\
        \hline
        AVE& 28 & 3,329 &402& 402&4/7 & 10 & 6/3 &140 \\
        AVK-100& 100 & 35,826 &11,955& 11,989&6/11 & 50 & 10/5 &1000 \\
        AVK-200& 200 & 68,320 & 22,798 & 22,882 &11/21 & 100 & 10/5 &2000 \\
        AVK-400& 400 & 140,497 & 46,885 & 47,045 &21/41 & 200 & 10/5 &4000 \\
      \shline
      \end{tabular}
      }
\end{table*}

  \begin{figure*}
    \centering
    \subfigure[AVE (3 phases)]{
    \includegraphics[width=4.2cm]{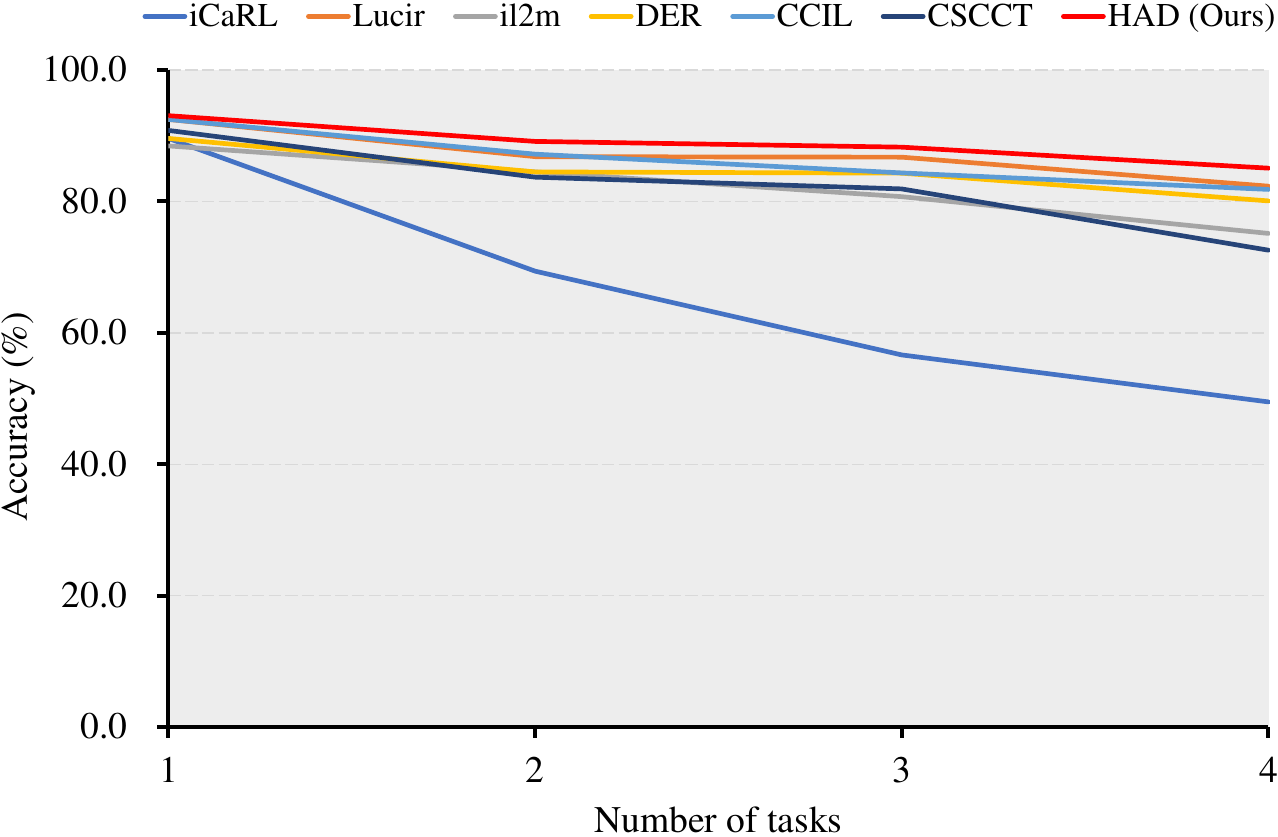}
    }
    \subfigure[AVK-100 (5 phases)]{
    \includegraphics[width=4.2cm]{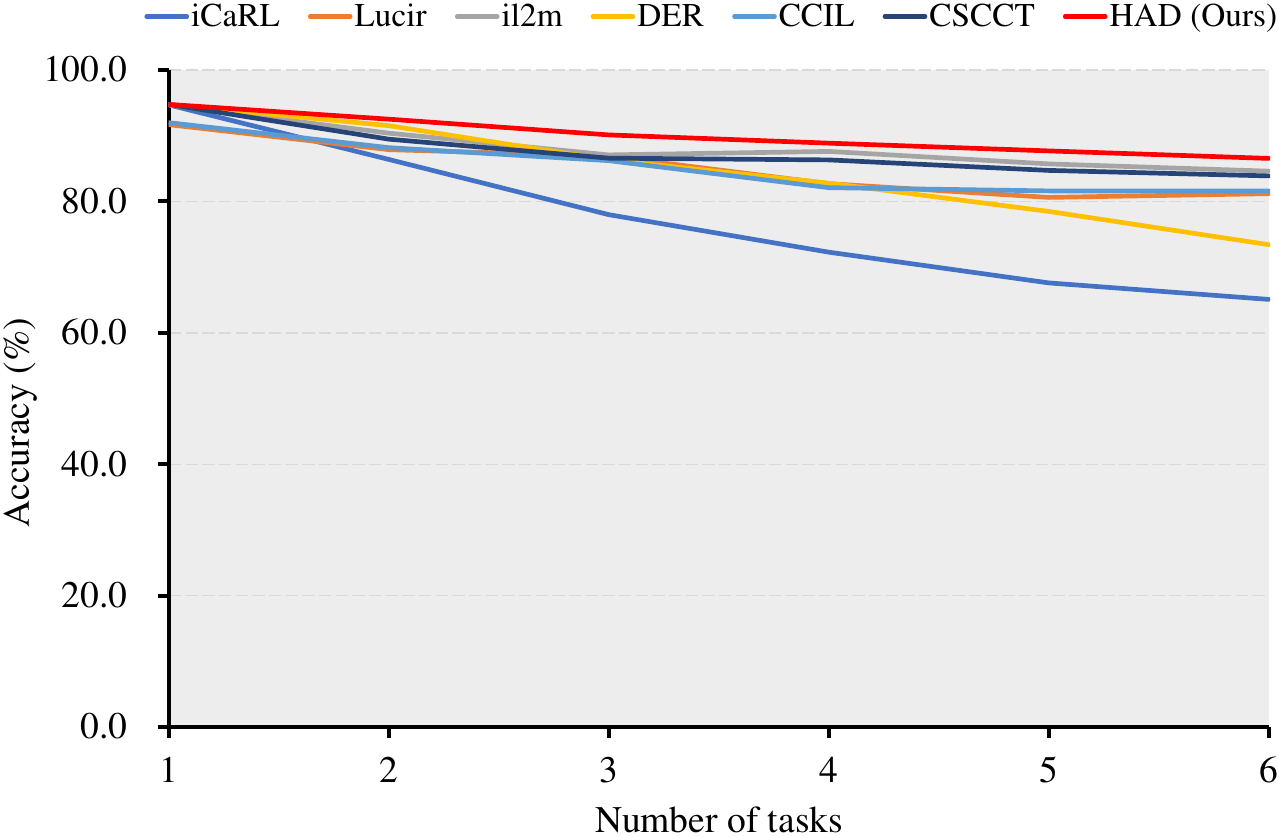}
    }
    \subfigure[AVK-200 (10 phases)]{
    \includegraphics[width=4.2cm]{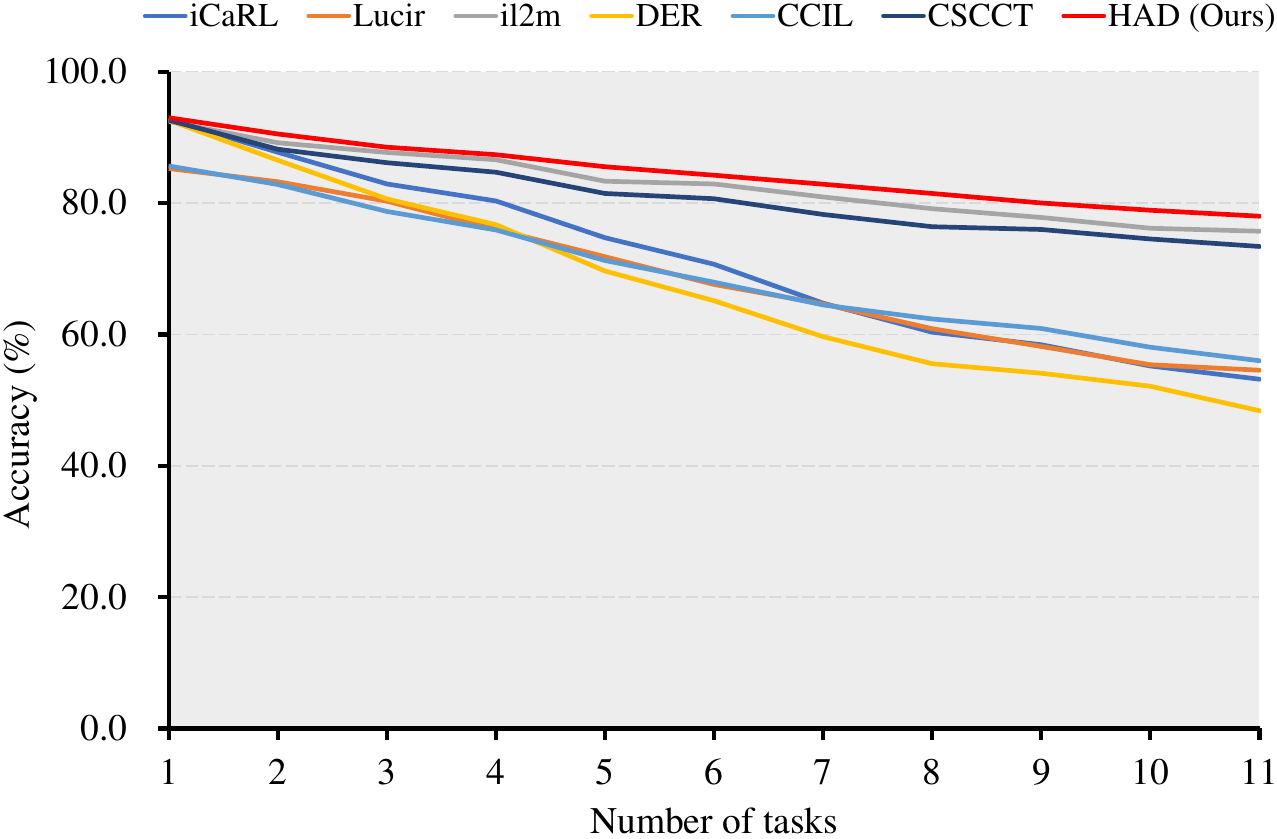}
    }
    \subfigure[AVK-400 (20 phases)]{
    \includegraphics[width=4.2cm]{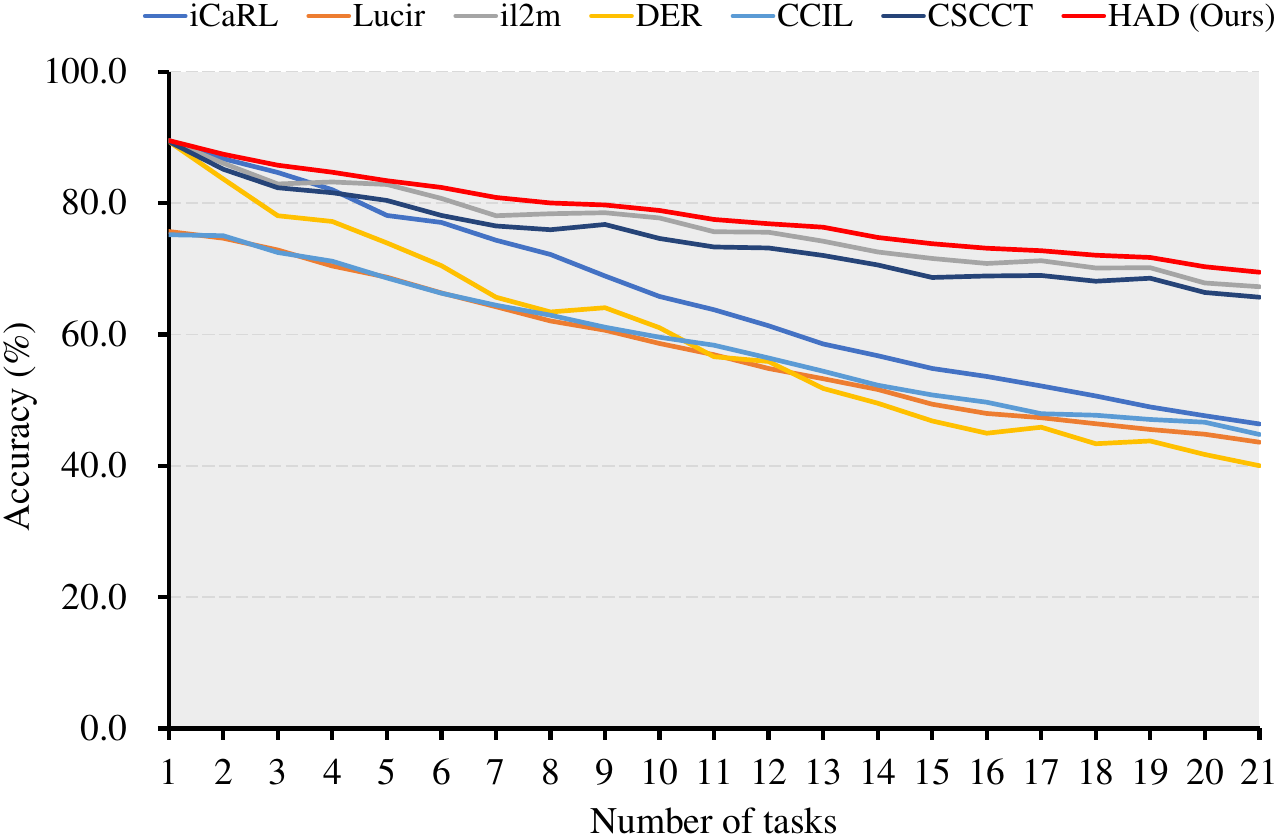}
    }
    \subfigure[AVE (6 phases)]{
    \includegraphics[width=4.2cm]{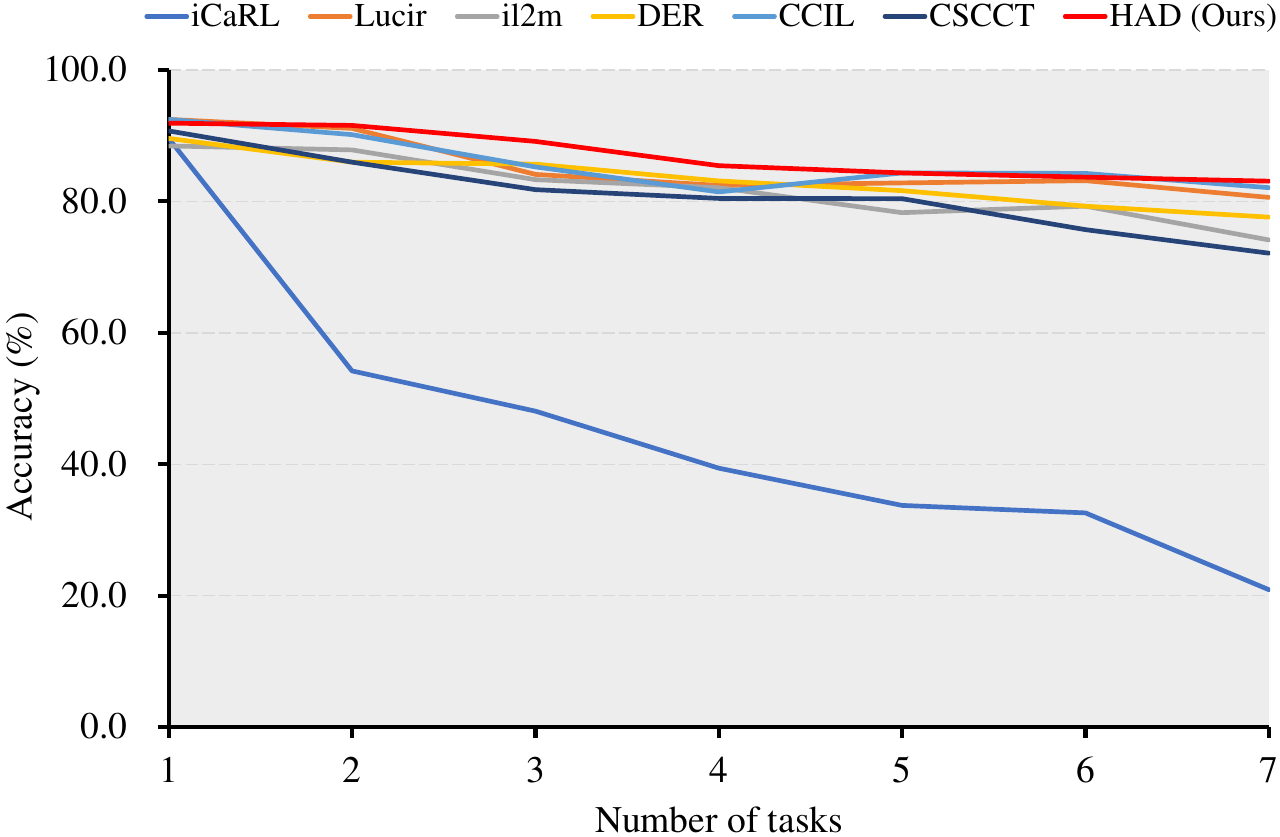}
        }
    \subfigure[AVK-100 (10 phases)]{
    \includegraphics[width=4.2cm]{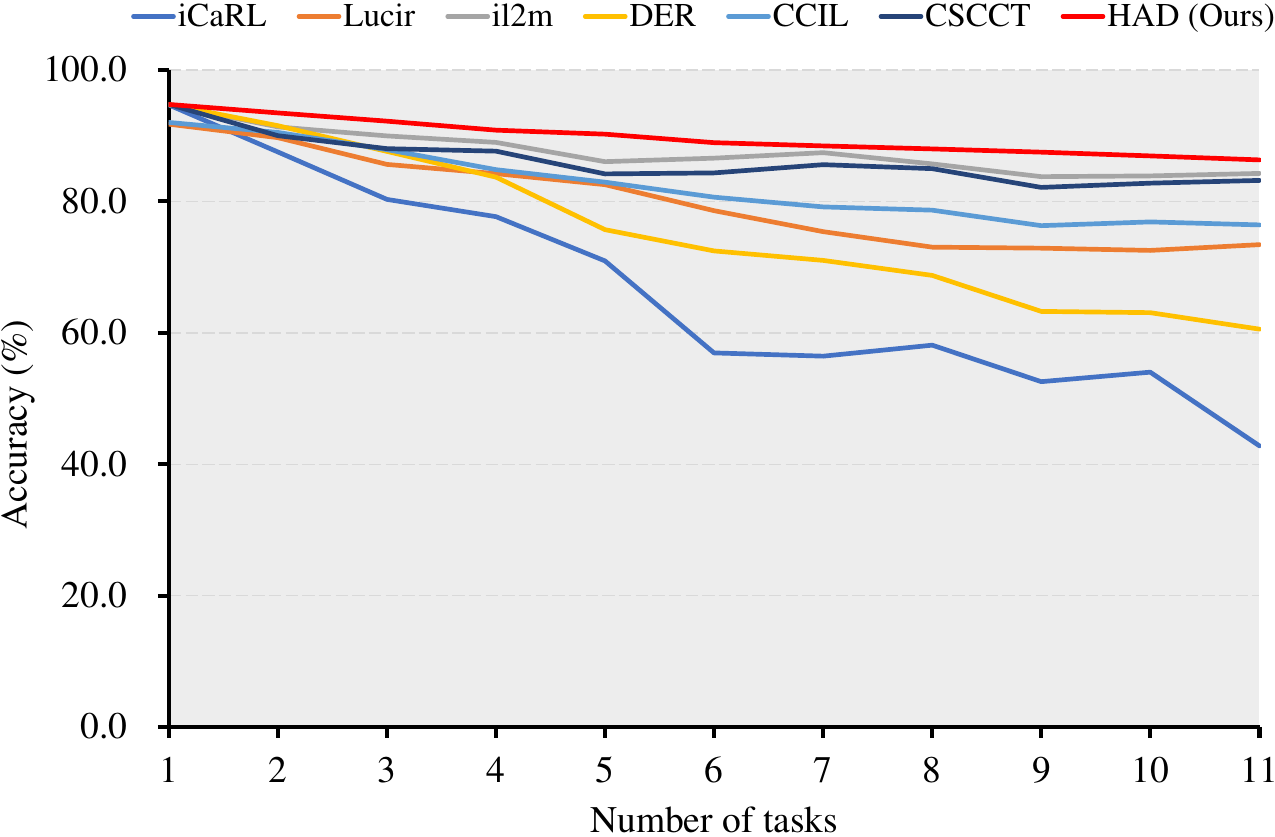}
    }
    \subfigure[AVK-200 (20 phases)]{
      \includegraphics[width=4.2cm]{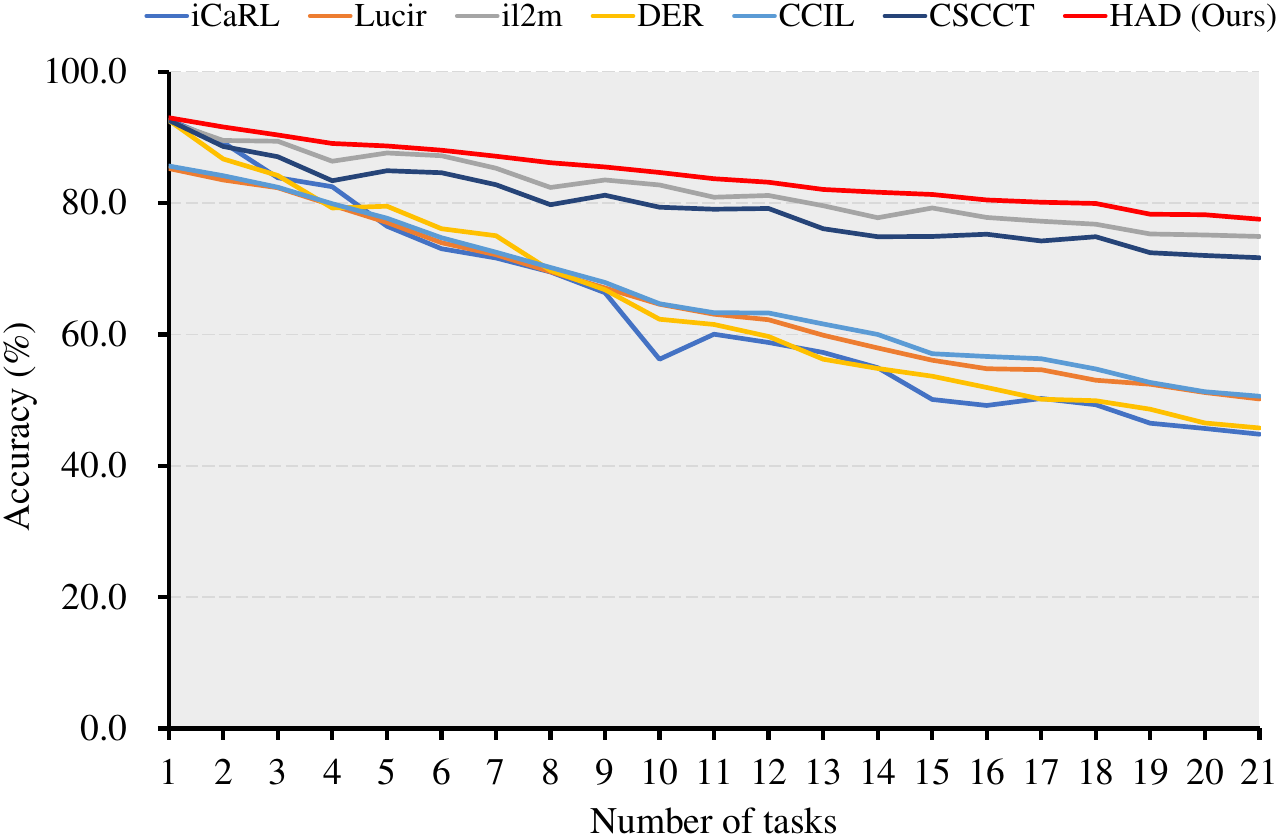}
      }
    \subfigure[AVK-400 (40 phases)]{
      \includegraphics[width=4.2cm]{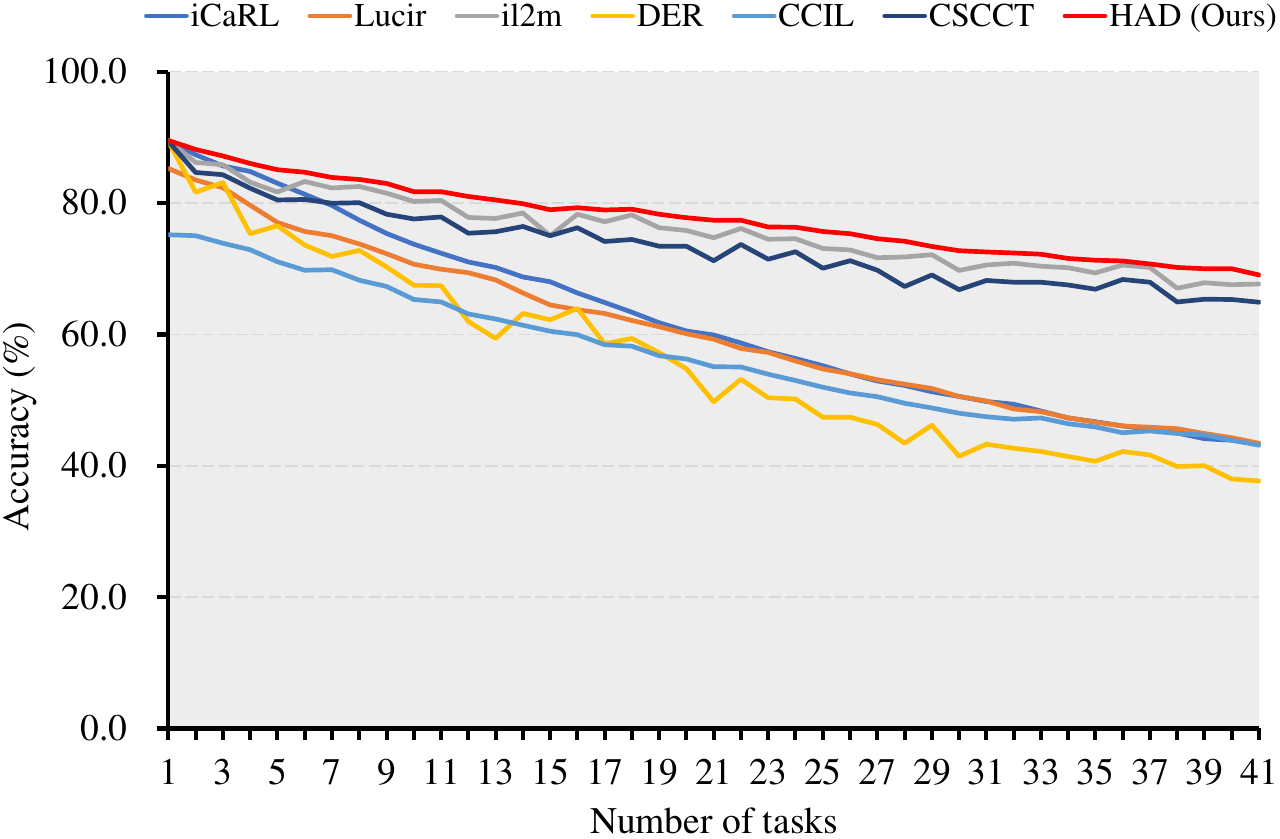}
      }
    \caption{The accuracy in AVE, AVK-100, AVK-200, and AVK-400 with different phases.}
  
    \label{main-results-fig}
  \end{figure*}

  \section{Experiments}
  \subsection{Training Details}
  \textbf{Datasets.} 
We use the AVE, AVK-100, AVK-200, and AVK-400 datasets for class-incremental audio-visual video recognition. The AVE dataset~\cite{a28}, derived from AudioSet~\cite{a45}, includes 4,143 videos across 28 categories, with 3,339 for training, 402 for validation, and 402 for evaluation. On the other hand, AVK-100, AVK-200, and AVK-400 are datasets specifically created for class-incremental audio-visual video recognition tasks, sourced from Kinetics-400~\cite{kay2017kinetics} without any additional videos. Due to some videos in Kinetics-400 having invalid YouTube download links and issues with extracting audio-visual features, we constructed the AVK-100, AVK-200, and AVK-400 datasets based on Kinetics-400 and divided them into training, validation, and evaluation sets ourselves.
AVK-100 contains 59,770 videos from 100 categories, with 35,826 for training, 11,955 for validation, and 11,989 for evaluation.
AVK-200 contains 114,000 videos from 200 categories, with 68,320 for training, 22,798 for validation, and 22,882 for evaluation.
AVK-400 includes 234,427 videos from 400 categories, with 140,497 for training, 46,885 for validation, and 47,045 for evaluation.

\noindent  \textbf{Benchmark Protocol.}
  We follow the standard protocol used in class incremental learning~\cite{a12, a13}, \emph{i.e.}, an initial base task followed by $N$ incremental tasks, and each task contains the same number of classes.
  For AVE, we select 10 classes as the initial base task, and divide the remaining 18 classes into 6/3 incremental tasks (6/3 phases), with each incremental task containing 3/6 classes.
  Similarly, for AVK-100, we choose 50 classes as the initial base task, and divide the remaining 50 classes into 5/10 incremental tasks (5/10 phases), where each incremental task contains 10/5 classes.
  For AVK-200, we select 100 classes as the initial base task, and divide the remaining 100 classes into 10/20 incremental tasks (10/20 phases), where each incremental task contains 10/5 classes.
  For AVK-400, we select 200 classes as the initial base task, and divide the remaining 200 classes into 20/40 incremental tasks (20/40 phases), where each incremental task contains 10/5 classes.
  We set the size of the memory bank as 140, 1000, 2000, and 4000 for AVE, AVK-100, AVK-200, and AVK-400, respectively.
  The primary dataset statistics and benchmark protocol are provided in Table~\ref{data_statistics}.

\begin{table*}
    \centering
    \scriptsize
    \setlength{\tabcolsep}{10.8pt}
    \footnotesize
    \renewcommand{\arraystretch}{1.2}
    \caption{Average Incremental Accuracy (AIA) / Final Incremental Accuracy (FIA) in AVE, AVK-100, AVK-200, and AVK-400 with different phases.}
    \label{main-results-tab}
    
    {
    \begin{tabular}{l  |c  c |c  c| c c|c c }
      \shline
      Methods & \multicolumn{2}{c|}{AVE}& \multicolumn{2}{c|} {AVK-100}& \multicolumn{2}{c|}{AVK-200}& \multicolumn{2}{c}{AVK-400}\\
      \hline
      No. of incremental tasks & 3 & 6 & 5 & 10 & 10 & 20  & 20 & 40\\
      \hline 
      Baseline         &74.8/57.2	&71.5/55.0&	58.6/37.5&	51.9/29.0&	41.5/17.3&	35.3/20.3 &	29.7/10.7 &	27.6/9.9  \\
      iCaRL \cite{a11} &66.3/49.5 	&45.5/20.9 &	77.4/65.1 &	66.6/42.8&	71.0/53.2 &	63.2/44.8  &	65.4/46.4 &	62.0/43.3 \\
      Lucir \cite{a12} &87.1/82.3 	&85.2/80.6 
        &	85.2/81.2 &	80.0/73.4 &	68.9/54.5 &	65.3/50.2  &	57.9/43.6&	55.8/43.4   \\
      il2m \cite{a6}   &82.2/75.1 	&81.9/74.1 &	88.4/84.6 &	87.5/84.3 &	82.9/75.7 &	82.0/74.9  &	76.4/67.2 &	75.7/67.7  \\
      DER \cite{a68} &84.6/80.1 	&83.3/77.6 &	84.6/73.4 &	75.7/60.6 &	67.4/48.4 &	64.3/45.8  &	59.4/40.0 &	56.0/37.7 \\
      CCIL \cite{a13}  &86.5/81.8 	&85.7/82.1 &	85.3/81.6 &	82.4/76.4 &	69.5/56.0 &	66.1/50.6  &	58.7/44.8 &	56.8/43.1  \\
      CSCCT \cite{a62}  &82.3/72.6 	&81.0/72.1 &	87.6/83.9 &	86.2/83.2 &	81.1/73.4 &	79.5/71.7  &	74.5/65.7  &	73.4/64.9  \\
      \hline
    HAD(Ours) &\textbf{88.9}/\textbf{85.1}	&\textbf{87.0}/\textbf{83.1}&	\textbf{90.1}/\textbf{86.6}&	\textbf{89.8}/\textbf{86.3}&	\textbf{84.6}/\textbf{78.0}&	\textbf{84.3}/\textbf{77.6} &	\textbf{78.2}/\textbf{69.5} &	\textbf{77.6}/\textbf{69.1}  \\
    \hline
    Joint-training &89.8	&89.8&	92.7&	92.7&	88.8&	88.8 &	84.8&	84.8 \\
  
    \shline
    \end{tabular}}
\end{table*}

\noindent  \textbf{Evaluation metric.}
  We adopt Average Incremental Accuracy (AIA)~\cite{a11,a86} as our evaluation metric, which represents the mean of the accuracies measured for already encountered data throughout all incremental phases (including the initial phase), serving as an indicator of the overall incremental effectiveness of the method when training on a sequence of tasks $\{\mathcal{T}_1,\mathcal{T}_2,\cdots, \mathcal{T}_S\}$:
  \begin{align}
    AIA = \frac{1}{S}\sum_{i=1}^{S}{IA}_i,
\end{align}
where Incremental Accuracy ${IA}_i$ represents the accuracy of the model on already encountered data after the completion of training task $\mathcal{T}_i$, respectively.  In the meanwhile, we also report Final Incremental Accuracy (FIA) result, which represents the accuracy of all the data at the final incremental phase:
  \begin{align}
    FIA = {IA}_s.
\end{align}

\noindent \textbf{Implementation Details}

  \begin{table}
    \centering
    \scriptsize
    \setlength{\tabcolsep}{15.0pt}
    \footnotesize
    \renewcommand{\arraystretch}{1.2}
    \caption{Component analysis of HAD framework on AVE 3 phases.}
    \label{ablation_1}
    
    {
      \begin{tabular}{c|c c c| c}
        \shline
        \multirow{2}{*}{Method} & \multirow{2}{*}{HAM} &\multirow{2}{*}{HLD} &\multirow{2}{*}{HCD} &\multirow{2}{*}{AIA} \\
        & & &  &  \\
        \hline
        Baseline&\xmark&\xmark&\xmark& 	74.8

        \\
        \hline
        HAM&\cmark&\xmark&\xmark& 	87.6 
        \\
        \hline

        \multirow{3}{*}{HDM}  &\xmark&\cmark&\xmark& 	80.5 
        \\
        &\xmark&\xmark& \cmark& 	78.2 
        \\
        &\xmark&\cmark&\cmark& 	82.2 
        \\
        \hline
        HAD &\cmark&\cmark&\cmark& 	\textbf{88.9} 
        \\
      \shline
      \end{tabular}
      }
  \end{table}

For each video, we sample frames at 8$fps$ and  divide the video into 10 non-overlapping snippets equally.
We utilize a frozen and pre-trained audio-visual embedding module to extract audio-visual features. Specifically, we employ the VGGish model~\cite{a45} for audio feature extraction, and the ResNet-152~\cite{a4} and 3D ResNet~\cite{a46} models for 2D and 3D visual feature extraction, respectively.
Audio features are extracted at the snippet level using the pre-trained VGGish model, while visual features are obtained by combining the outputs of ResNet-152 and 3D ResNet to create fused visual snippet-level features for each video snippet.  The ResNet-152 model is pre-trained on the ImageNet dataset~\cite{5206848}, and the 3D ResNet model is pre-trained on the Kinetics-400 dataset~\cite{kay2017kinetics}. Similarly, the VGGish model is pre-trained on the Audio-Set dataset~\cite{7952261}.
Given that the AVE dataset is a subset of the Audio-Set dataset, and the AVK100, AVK200, and AVK400 datasets are subsets of the Kinetics-400 dataset, the pre-trained audio-visual embedding module provides effective feature representations  for subsequent fusion and classification modules.
  In the modal fusion network, we use a hybrid attention network~\cite{a29} to obtain the fused features.
  For the classifier, we adopt a cosine-normalized last layer similar to CCIL~\cite{a13}, \emph{i.e.}, calculating the cosine similarity between the normalized features and normalized class-weight vectors for AVE.
  For AVK-100, AVK-200, and AVK-400, we utilize the last linear layer for classification.
  Similar to CCIL~\cite{a13}, the exemplar set also includes an equal size of exemplar samples from current classes.
  The model is trained with Adam~\cite{a47} optimizer with the learning rate of 3e-5 and epochs of 10.
  We set $\lambda = 0.05$, $\beta = 5$, $\gamma = 0.2$, and $\eta = 25$.  
  Batch sizes are 16 and 256 for AVE and AVK-100/200, respectively. 
  The  code of the proposed  method is available at \textcolor{magenta}{https://github.com/Play-in-bush/HAD}.

  \begin{table*}
    \begin{minipage}{0.33\textwidth}
    \caption{Hierarchy in HAM module on AVE 3 phases.}
    \scriptsize
    \centering
    \setlength{\tabcolsep}{11.4pt}
    \footnotesize
    \renewcommand{\arraystretch}{1.2}
    \begin{tabular}{c|c c|c}
      \shline
      \multirow{2}{*}{Method} & \multirow{2}{*}{LMA} &\multirow{2}{*}{HVA} &\multirow{2}{*}{AIA} \\
      & & & \\
      \hline
      Baseline &\xmark&\xmark& 74.8\\
      \hline
  
      \multirow{3}{*}{HAM}  &\cmark&\xmark&	86.4   \\
   &\xmark& \cmark&	87.0  \\
    &\cmark&\cmark&	\textbf{87.6}\\
    \shline
    \end{tabular}
    \label{ablation_2}
    \end{minipage}
    \begin{minipage}{0.33\textwidth}
    \caption{Hierarchy in  HLD module on AVE 3 phases.}
    \scriptsize
    \centering
    \setlength{\tabcolsep}{11.4pt}
    \footnotesize
    \renewcommand{\arraystretch}{1.2}
    \begin{tabular}{c|c c|c}
      \shline
  
      \multirow{2}{*}{Method} & \multirow{2}{*}{SLD} &\multirow{2}{*}{DLD} &\multirow{2}{*}{AIA} \\
      & & & \\
      \hline
      Baseline &\xmark&\xmark& 74.8\\
      \hline
      \multirow{3}{*}{HLD}  &\cmark&\xmark&	80.3   \\
   &\xmark& \cmark&	79.7  \\
    &\cmark&\cmark&	\textbf{80.5}\\
    \shline
    \end{tabular}
    \label{ablation_3}
    \end{minipage}
    \begin{minipage}{0.33\textwidth}
      \caption{Hierarchy in HCD module on AVE 3 phases.}
      \scriptsize
      \centering
      \setlength{\tabcolsep}{11.4pt}
      \footnotesize
      \renewcommand{\arraystretch}{1.2}
      \begin{tabular}{c|c c|c}
        \shline
    
        \multirow{2}{*}{Method} & \multirow{2}{*}{SCD} &\multirow{2}{*}{VCD} &\multirow{2}{*}{AIA} \\
        & & & \\
        \hline
        Baseline &\xmark&\xmark& 74.8\\
        \hline
        \multirow{3}{*}{HCD}  &\cmark&\xmark&	76.9   \\
        &\xmark& \cmark&	77.9  \\
         &\cmark&\cmark&	\textbf{78.2}\\
      \shline
      \end{tabular}
      \label{ablation_4}
      \end{minipage}
    \end{table*}

  \subsection{Comparison with Existing Methods}
  In this section, we compare the proposed method with classical exemplar-based methods, such as iCaRL~\cite{a11}, Lucir~\cite{a12}, il2m~\cite{a6}, DER\cite{a68},  CCIL~\cite{a13}, and CSCCT~\cite{a62}.
  To ensure a fair comparison, all these methods are evaluated on the same dataset division and leverage both audio and visual features of the videos. Each method is re-implemented using the same audio and visual features generated by the pre-trained audio-visual embedding module. 
  Moreover, a random sampling strategy is employed to construct the exemplar set for each method.
  The comparative results are shown in Table~\ref{main-results-tab} and Figure~\ref{main-results-fig}.
  The term `Baseline' represents using the fine-tune strategy to infer the model on each incoming dataset with the classifier used in our method. Specifically, `Baseline' only concentrates on the audio-visual video recognition of current task by performing classification loss for current task data, and past task data are not available for augmentation or distillation. 
  As the number of tasks increases, the performance of all methods declines overall, which demonstrates the necessity of class incremental audio-visual video recognition task.
 Compared with the ‘Baseline’, HAD achieves improvements of 14.1\%/27.9\% (15.5\%/28.1\%), 31.5\%/49.1\% (37.9\%/57.3\%), 43.1\%/60.7\% (49.0\%/57.3\%), and 48.5\%/58.8\% (50.0\%/59.2\%) about Average Incremental Accuracy / Final Incremental Accuracy  metrics in AVE 3 phases (6 phases), AVK-100 5 phases (10 phases), AVK-200 10 phases (20 phases) and AVK-400 20 phases (40 phases), respectively.

  Additionally, HAD outperforms existing methods on all datasets,  \emph{e.g.}, obtaining  Average Incremental Accuracy / Final Incremental Accuracy  of 88.9\%/85.1\% (87.0\%/83.1\%), 90.1\%/86.6\% (89.8\%/86.3\%), 84.6\%/78.0\% (84.3\%/77.6\%) and 78.2\%/69.5\% (77.6\%/69.1\%) in AVE 3 phases (6 phases), AVK-100 5 phases (10 phases), AVK-200 10 phases (20 phases) and AVK-400 20 phases (40 phases), respectively. 
  Moreover, as illustrated in Figure~\ref{main-results-fig}, HAD surpasses other methods in nearly all incremental phases, demonstrating its superiority and stability.
  From Table~\ref{main-results-tab}, we also observe that existing method `il2m' yields favorable results in large dataset, \emph{e.g.},  88.4\%/84.6\% (87.5\%/84.3\%), 82.9\%/75.7\% (82.0\%/74.9\%), and 76.4\%/67.2\% (75.7\%/67.7\%) about Average Incremental Accuracy / Final Incremental Accuracy  in AVK-100 5 phases (10 phases), AVK-200 10 phases (20 phases) and AVK-400 20 phases (40 phases), respectively. 
  This outcome is attributed to the method's storage of exemplar data and statistics of old classes.
  In contrast to `il2m', HAD only stores exemplar data for knowledge preservation.
  Despite storing fewer exemplars, HAD still outperforms `il2m' across all four datasets.
  The superior results confirm the effectiveness of the proposed HAD.

  \subsection{Ablation Study}
  \subsubsection{Elements of HAD}
  We analyze the role of HAM, HDM, HLD, and HCD within the proposed HAD framework based on the setting of the AVE 3 phases with Average Incremental Accuracy (AIA) metric.
  Table~\ref{ablation_1} shows that using only HAM yields a 12.8\% improvement compared with the baseline, indicating that applying hierarchical feature augmentation to exemplar data enhances knowledge preservation. 
  Using HLD and HCD improves the baseline by 5.7\% and 3.4\%, respectively, demonstrating that both logical distillation and correlative distillation are helpful for model knowledge retention.
  The  comparison of HAD, HLD, and HCD (87.6\% v.s 80.5\% v.s 78.2\%) reveals that the hierarchical augmentation module effectively mitigates catastrophic forgetting in class incremental audio-visual video recognition, with hierarchical logical distillation slightly outperforming hierarchical correlative distillation. 
  By combining HLD and HCD, HDM obtains an average improvement of 7.4\%, proving that logical distillation and correlative distillation complement each other.
  By combining HAM and HDM, HAD improves Average Incremental Accuracy (AIA) by 1.3\% and 6.7\% for HAM and HDM, respectively.
  The superior performance indicates that data knowledge preservation and model knowledge preservation are essential for class incremental audio-visual video recognition.

  \subsubsection{Effect of Hierarchical Structure}
  We analyze the necessity of hierarchical structure in feature  augmentation, \emph{i.e.,} logical distillation and correlative distillation with Average Incremental Accuracy (AIA) metric, and summarize the  results in Table~\ref{ablation_2}, Table~\ref{ablation_3}, and Table~\ref{ablation_4}, respectively.
  
  Table~\ref{ablation_2} demonstrates that employing the low-level modal augmentation (LMA) and high-level video augmentation (HVA) both result in  higher performance compared with the baseline.  For example,  Average Incremental Accuracy (AIA) is improved from 74.8\% to 86.4\%/87.0\% for LMA/HVA, indicating that both low-level feature augmentation and high-level feature augmentation effectively maintain knowledge of previous classes.
  By combining LMA and HVA,  HAM obtains an Average Incremental Accuracy (AIA) of 87.6\%, suggesting that low-level modal augmentation and high-level video augmentation reinforce each other. 
  Consequently, it is necessary to consider both low-level and high-level feature augmentations.
  
  Table~\ref{ablation_3} shows that using the video-level logical distillations (SLD) and distribution-level logical distillations (DLD) yields improvements of 5.5\% and 4.9\%  compared with the baseline, respectively, illustrating that both video-level logical distillation and distribution-level logical distillation effectively preserve model knowledge.
  Combining SLD and DLD achieves the best Average Incremental Accuracy (AIA) of 80.5\%, verifying the necessity of hierarchical logical distillation.
  
  Table~\ref{ablation_4} indicates that employing the snippet-level correlative distillation (SCD) and video-level correlative distillation (VCD)  leads to improvements of 2.1\% and 3.1\% over the baseline, respectively, demonstrating the effectiveness of snippet-level correlative distillation and video-level correlative distillation. 
  HCD performs better by combining SCD and VCD, supporting the rationale behind hierarchical distillation.
  
  From Table~\ref{ablation_2}, Table~\ref{ablation_3}, and Table~\ref{ablation_4}, we can conclude that it is necessary to consider the hierarchical structure of the model and data to preserve data knowledge and model knowledge for class incremental audio-visual video recognition.

  \begin{figure}
    \centering
      \subfigure[]{
        \includegraphics[width=4cm]{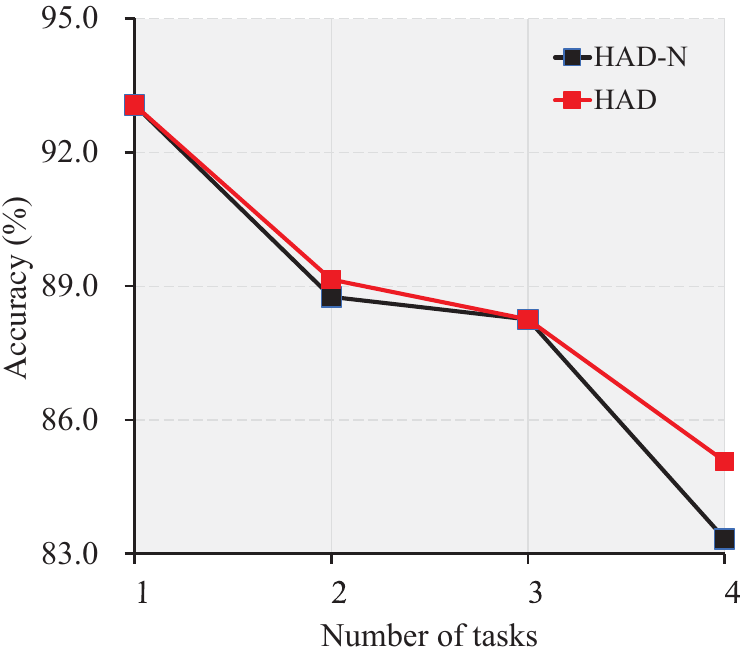}
        }
      \subfigure[]{
        \includegraphics[width=3.8cm]{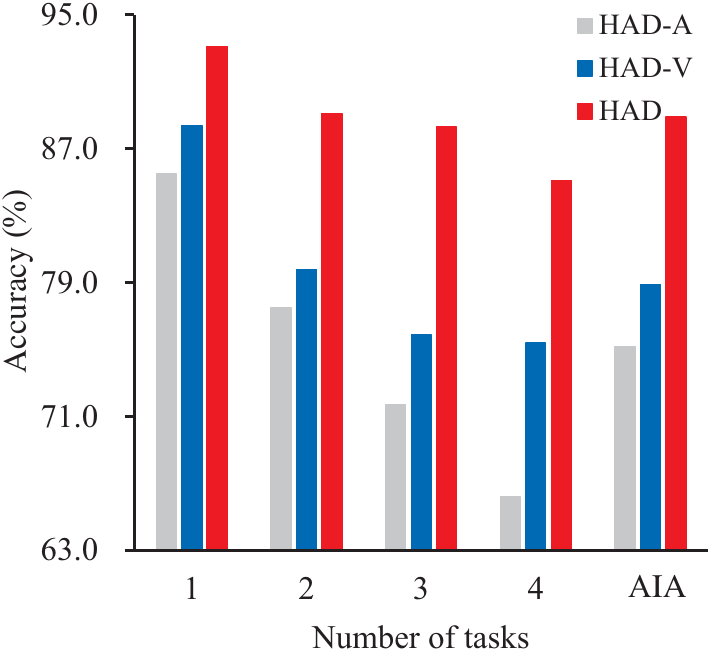}
        }

    \caption{\footnotesize{(a) Analysis of augmentation noise on AVE 3 phases. (b) Analysis of multimodal on AVE 3 phases.}}
    
    \label{ablation_5}
    \end{figure}

  \subsubsection{Analysis of Augmentation Noise}
  To illustrate why the low-level and high-level feature augmentation should impact the parameters of different modules, we analyze the effect of error information accumulation caused by different augmentations, shown in Figure~\ref{ablation_5}(a).
  HAD-N updates the parameters of the audio-visual fusion module $F$ and classifier $C$ with low-level feature augmentation.
  The low-level feature augmentation in HAD only updates the audio-visual fusion module $F$.
  As depicted in Figure~\ref{ablation_5}(a), HAD-N achieves a lower performance than HAD, indicating that the error caused by low-level feature augmentation can degrade the classifier $C$.
  Therefore, it is reasonable to conduct the low-level feature augmentation and high-level video augmentation for adjusting the parameters of audio-visual fusion module $F$ and classifier $C$, respectively,
  The proposed method not only takes full advantage of the generalization provided by low-level and high-level feature augmentation but also avoids the accumulation of errors caused by feature augmentation.

\subsubsection{Analysis of Multi-modal}
To verify the necessity of using audio and video multi-modal information for CIAVVR, we illustrate how class incremental learning performance changes when using audio information (HAD-A) or visual information (HAD-V) solely in Figure~\ref{ablation_5}(b).
We observe that the results for all tasks in HAD-A and HAD-V are lower than those of HAD, demonstrating that HAD, which exploits multi-modal information, achieves better performance in video-level class incremental learning than using only single-model information.
Furthermore, we observe that the performance gap between HAD-A/HAD-V and HAD in the 4-th task (18.9\%/9.7\%) is larger than that in the first task (7.6\%/4.7\%), illustrating that  utilizing multi-modal information suffers  less from catastrophic forgetting in video-level class incremental learning.
Moreover, HAD, which fuses audio and visual information, outperforms HAD-A and HAD-V by 13.7\% and 10.0\% on Average Incremental Accuracy (AIA). 
This shows that the audio information complements visual information, and fusing the audio and visual information achieves better performance. The above results demonstrate the necessity of integrating multi-modal information for video recognition tasks.

  \begin{figure*}
    \centering

    \subfigure[$\beta$]{
    \includegraphics[width=5.1cm]{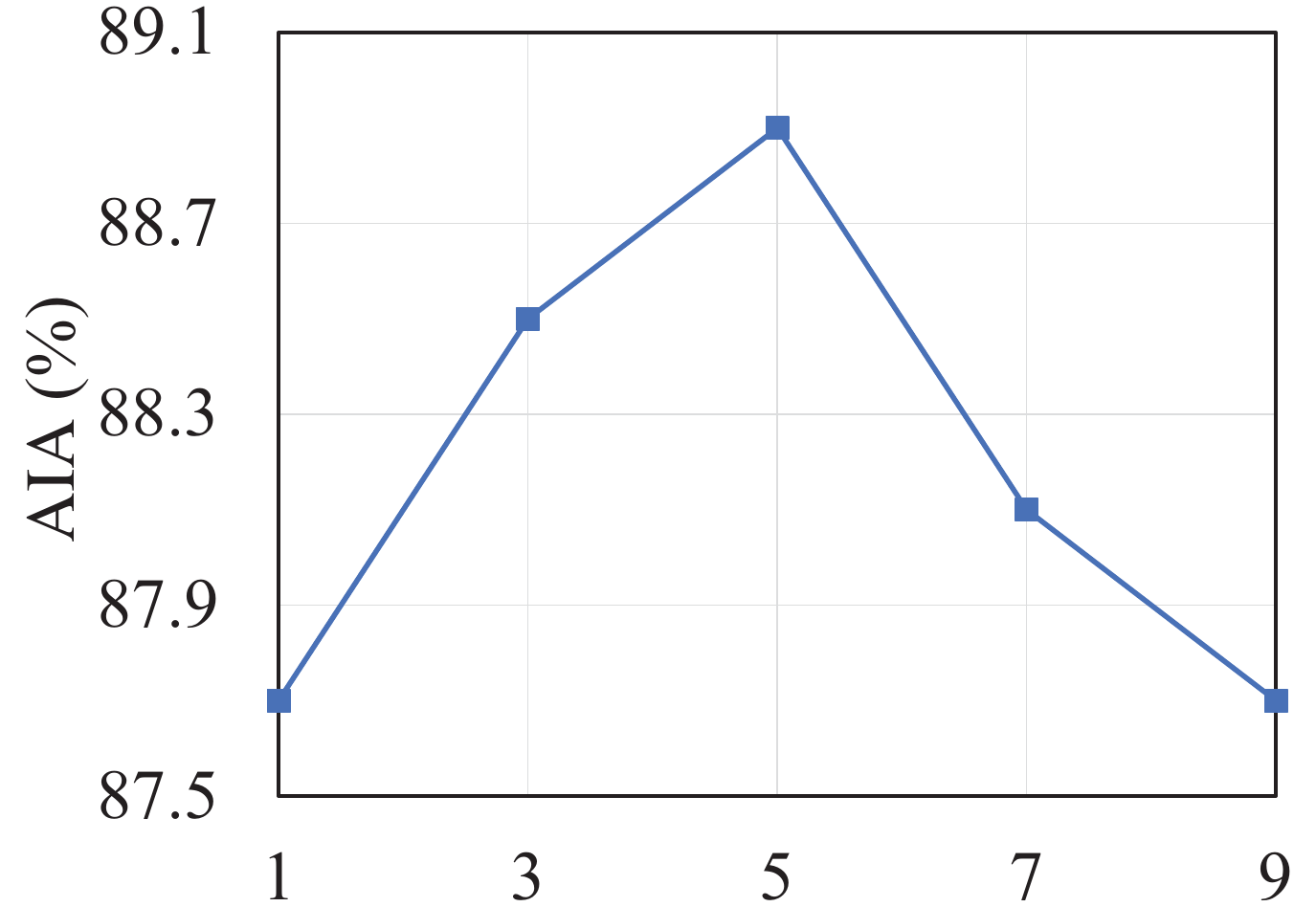}
    }
    \subfigure[$\gamma$]{
    \includegraphics[width=5.1cm]{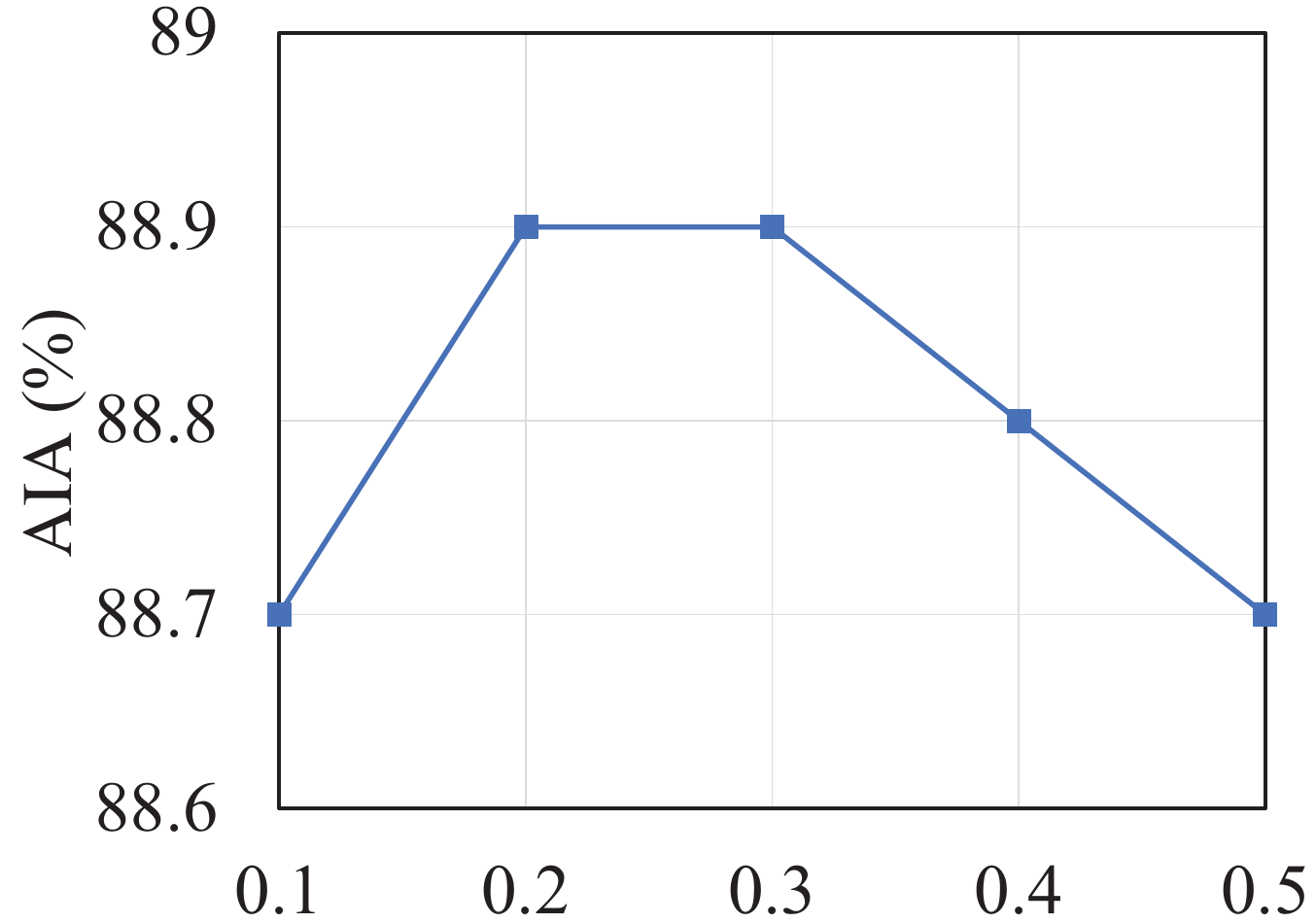}
    }
    \subfigure[$\eta$]{
      \includegraphics[width=5.1cm]{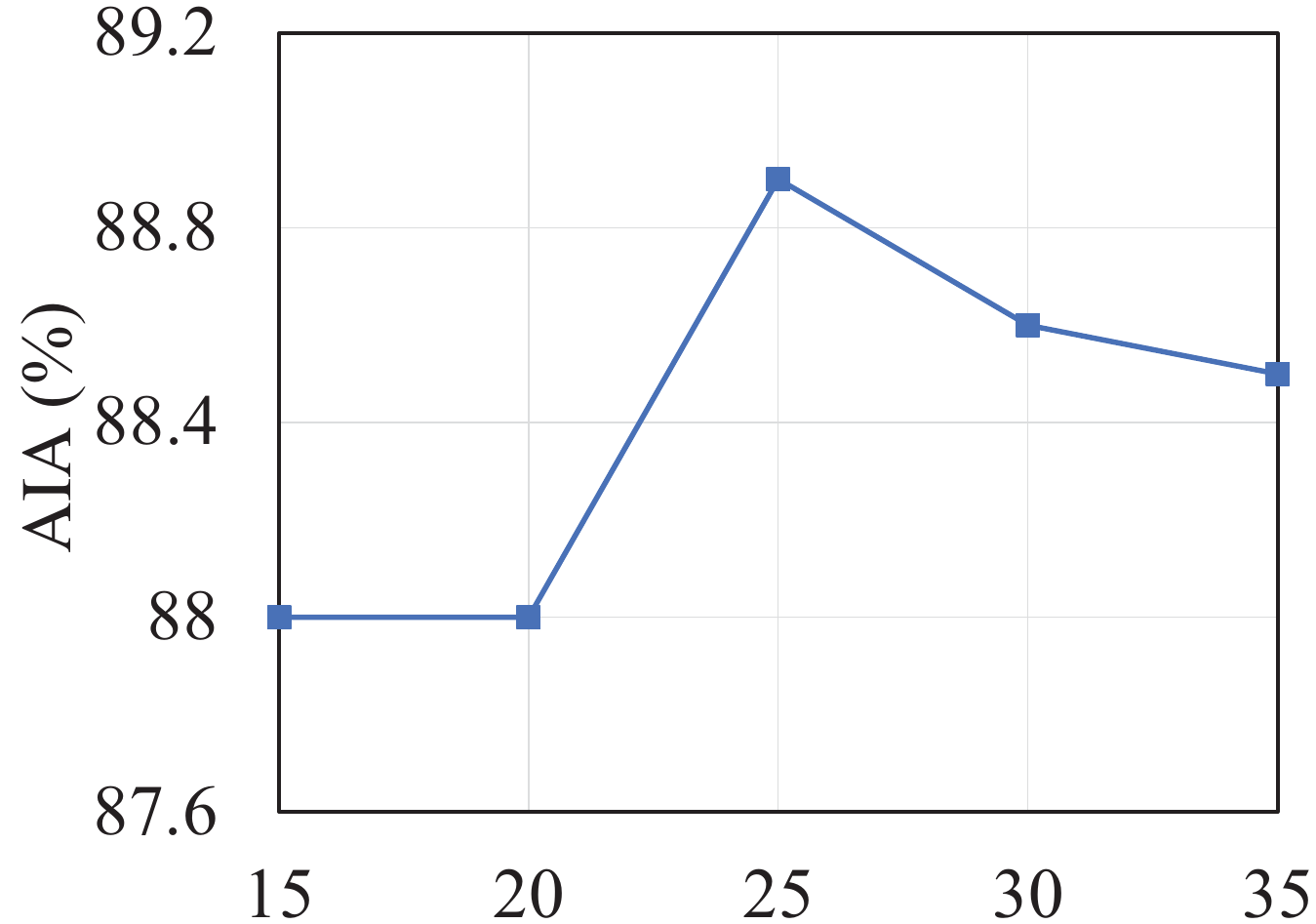}
      }
  
    \caption{Plot of various hyperparameter analyses on AVE 3 phases.
    (a) Varying the trade-off parameter $\beta$ for the loss of hierarchical augmentation module. 
    (b) Varying the trade-off parameter $\gamma$ for  the loss of hierarchical logical distillation in hierarchical distillation module.  (c)  Varying the trade-off parameter $\eta$ for  the loss of hierarchical correlative distillation in hierarchical distillation module.}
    \label{hyperparameter}
  \end{figure*}

  \begin{figure}
    \begin{center}
    {\includegraphics[width=0.75\linewidth]{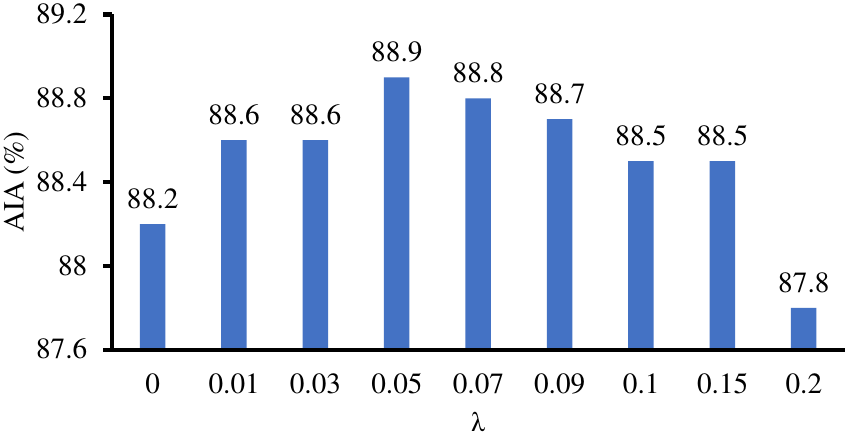}}
    \end{center}
    \caption{Sensitive analysis about intensity of Gaussian augmentation $\lambda$ on AVE 3 phases.
    }\label{sensitiveanalysis}
    \end{figure}

\subsubsection{Sensitive analysis about intensity of Gaussian augmentation $\lambda$}
Because Gaussian augmentation is incorporated into both the model's low-level modal features and high-level video features, the Gaussian noise parameter $\lambda$ is critical as it determines the balance between enhancing generalization and the risk of introducing harmful noise. Our results, shown in Figure~\ref{sensitiveanalysis}, reveal that increasing $\lambda$  from 0 to 0.05 leads to a slight improvement in the model's accuracy, from 88.2\% to 88.9\%. This suggests that a small amount of Gaussian noise can actually improve generalization without negatively affecting the model's predictions. The performance stabilizes between 88.6\% and 88.7\% when $\lambda$  ranges from 0.01 to 0.09, indicating an optimal noise level for maximum robustness. Furthermore, the best performance of the model, at 88.9\%, is observed when $\lambda$  is specifically at 0.05. However, when $\lambda$  increases to 0.15 and further to 0.2, there's a decline in accuracy to 88.5\% and 87.8\%, respectively, signifying that the negative effects of noise begin to outweigh its benefits. These results indicate that Gaussian noise can enhance model generalization within a certain range. Within this range, the model demonstrates good tolerance and robustness to noise.

\subsubsection{Hyperparameter Analysis}
We conduct hyperparameter analyses with Average Incremental Accuracy (AIA) metric and summarize the related results in Figure~\ref{hyperparameter}.
Figure~\ref{hyperparameter}(a) indicates that the trade-off parameter $\beta = 5$ for the loss of hierarchical augmentation module outperforms other settings.
Reducing $\beta$ causes the model to retain  less knowledge of old classes, resulting in inadequate solutions for catastrophic forgetting.
Furthermore, increasing $\beta$ constrains the the model to focus more on the knowledge of old classes, limiting its ability to learn new class knowledge. 
Figure~\ref{hyperparameter}(b) and Figure~\ref{hyperparameter}(c) suggest that  proper  trade-off parameters $\gamma = 0.2$ and $\eta = 25$ are crucial for balancing  old model knowledge preservation and current model knowledge learning.
Remembering the old model knowledge can overcome catastrophic forgetting of old classes, but also can restrict the learning of current classes.

  \begin{figure}
    \begin{center}
    {\includegraphics[width=0.65\linewidth]{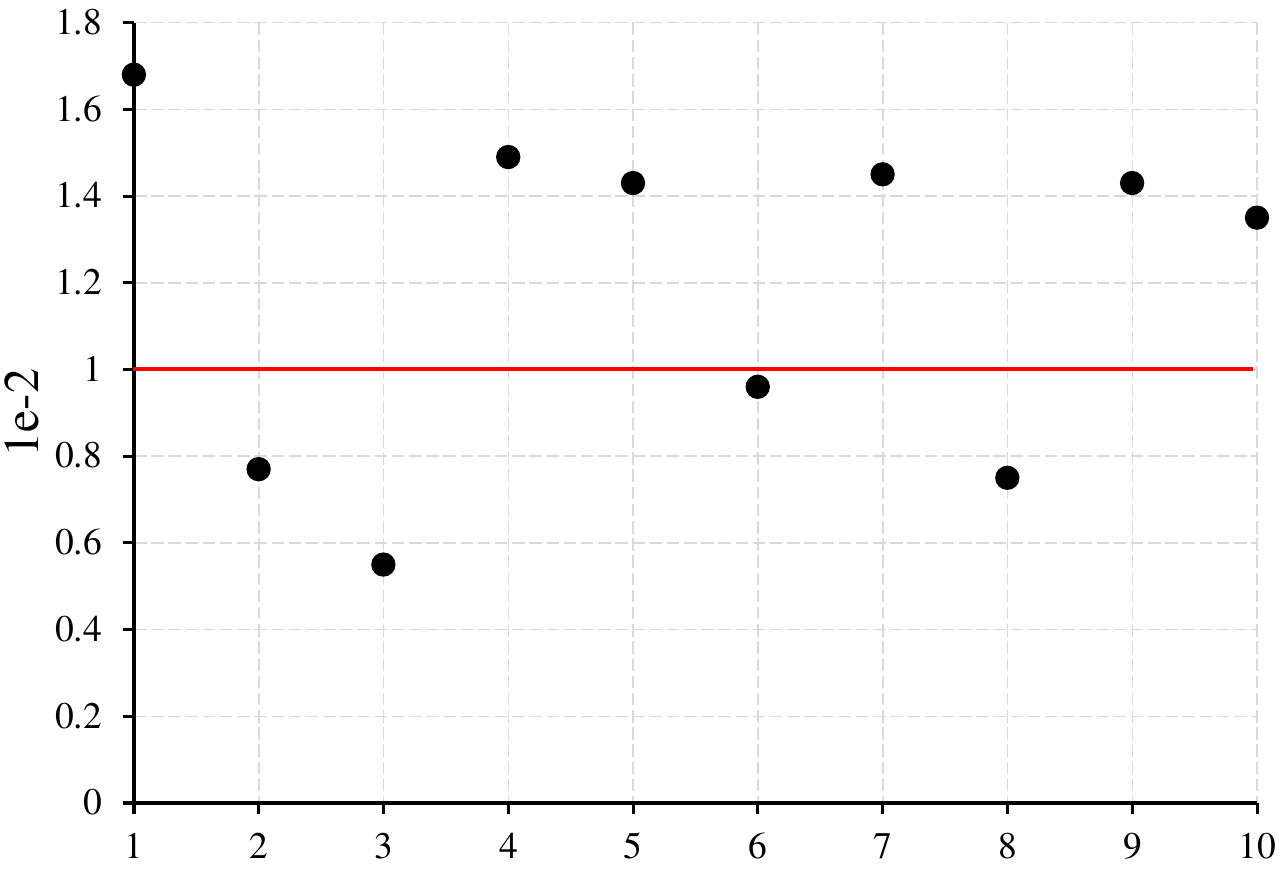}}
    \end{center}
    \caption{The distance between $\mathcal{F}(\mathcal{T}'_s)$ and $\mathcal{F}(\mathcal{T}_s)$.
    }\label{supp}
    \end{figure}

\subsubsection{1-Lipschitz Continuous in  $\mathcal{F}$}
In the theoretical analysis of hierarchical augmentation, we assume that the audio-visual fusion module $\mathcal{F}$ does not satisfy the 1-lipschitz continuous, and  $\mathcal{F}(\mathcal{T}')$ is prone to deviating from the  distribution of $\mathcal{F}(\mathcal{\mathcal{T}})$. 
We thus conduct an experiment to verify this assumption.
We randomly sample ten low-level modal features $\mathcal{T}_s$ from $\mathcal{T}$, and  add noise with a distance of 1e-2 to the low-level modal features $\mathcal{T}_s$, \emph{i.e.}, $\mathcal{T}'_s = \mathcal{T}_s +$ 1e-2. 
Finally, we calculate the distance between $\mathcal{F}(\mathcal{T}'_s)$ and $\mathcal{F}(\mathcal{T}_s)$.
From Figure~\ref{supp}, we can see that the distance  between $\mathcal{F}(\mathcal{T}'_s)$ and $\mathcal{F}(\mathcal{T}_s)$ of most samples is greater than 0.01, which violates the 1-Lipschitz continuous.
Furthermore, the mean distance between $\mathcal{F}(\mathcal{T}'_s)$ and  $\mathcal{F}(\mathcal{T}_s)$ is 1.2e-2, demonstrating that $\mathcal{F}(\mathcal{T}')$ is prone to deviating from the  distribution of $\mathcal{F}(\mathcal{\mathcal{T}})$.
The above results verify the rationality of our assumption.

\subsubsection{Analysis of out-of-distribution dataset}

In the main experiments, the AVE dataset is a subset of the Audio-set, which is a pre-trained dataset for the audio feature extraction network VGGish. AVK100, AVK200, and AVK400 are subsets of the Kinetics-400 dataset for the 3D ResNet visual feature extraction network. Therefore, we can obtain effective feature representations from these datasets to provide good inputs for subsequent audio-visual fusion and classification modules. However, for audio-visual incremental learning tasks, the dataset may not always align with the pre-trained model's dataset, potentially resulting in suboptimal or corrupted features. To assess our model's generalization capabilities for out-of-distribution datasets, we introduce a new dataset called Extra-AVK. This dataset consists of 200 classes selected from Kinetics-600, which are distinct from the Kinetics-400 classes. We extract 200 videos per class and split them into training, validation, and evaluation sets with a 6:2:2 ratio, resulting in 120 training, 40 validation, and 40 test videos per class. Unlike the other datasets, the Extra-AVK dataset does not have pre-trained information for the audio, 2D visual, and 3D visual networks. We use 100 classes from Extra-AVK as the initial base task and divide the remaining 100 classes into 10 or 20 incremental tasks (phases), with each task containing 10 or 5 classes, respectively.

From the Table~\ref{Extra-AVK}, it's evident that our method HAD outperforms other approaches on the Extra-AVK dataset, which is completely new to the audio, 2D visual, and 3D visual networks. 
For example, HAD achieves the best Average Incremental Accuracy / Final Incremental Accuracy of 58.6\%/45.5\% (57.0\%/44.6\%) on Extra-AVK with 10 phases (20 phases), underscoring our method's effectiveness and generalization capability. 
Moreover, we observe a significant decline in the performance of Joint-training and Baseline on Extra-AVK compared to their performance on AVK-200 in Table~\ref{main-results-tab}. Specifically, Joint-training on Extra-AVK sees a 26.0\% drop compared to AVK-200. The Baseline method on Extra-AVK with 10 phases (20 phases) shows a decrease in Average Incremental Accuracy / Final Incremental Accuracy by 27.9\%/13.4\% (25.9\%/18.1\%) compared to AVK-200 with 10 phases (20 phases).
This decline indicates that the pre-trained audio-visual embedding module lacks sufficient prior knowledge for the Extra-AVK dataset, leading to inferior feature extraction and, consequently, poorer audio-visual classification results. This suggests that for future audio-visual incremental learning tasks, employing more advanced audio-visual embedding module could enhance performance.

\begin{table}
\centering
\scriptsize
\setlength{\tabcolsep}{18.8pt}
\footnotesize
\renewcommand{\arraystretch}{1.2}
\caption{Average Incremental Accuracy (AIA) / Final Incremental Accuracy (FIA) in Extra-AVK with 10/20 phases.}
\label{Extra-AVK}
{
\begin{tabular}{l  |c  c }
  \shline
  Methods & \multicolumn{2}{c}{Extra-AVK}\\
  \hline
  No. of incremental tasks & 10 & 20 \\
  \hline 
  Baseline         & 13.6 / 3.9 & 9.4 / 2.2 \\
  iCaRL \cite{a11} & 51.7 / 37.1 & 47.7 / 28.6 \\
  Lucir \cite{a12} & 33.0 / 22.8 & 31.6 / 22.0 \\
  il2m \cite{a6}   & 56.0 / 42.5 & 53.7 / 39.8 \\
  DER \cite{a68}   & 43.1 / 25.4 & 38.3 / 25.0 \\
  CCIL \cite{a13}  & 36.7 / 29.4 & 33.6 / 26.6 \\
  CSCCT \cite{a62} & 55.4 / 42.5 & 51.6 / 37.8 \\
  \hline
HAD(Ours) & \textbf{58.6} / \textbf{45.5} & \textbf{57.0} / \textbf{44.6} \\
\hline
Joint-training & 62.8 & 62.8 \\
\shline
\end{tabular}}
\end{table}

\subsubsection{Compared with large-scale fully-supervised method} 
To illustrate the rationality of the joint-training (upper-bound) results in our model framework, we compare the fully supervised performance of our model with that of state-of-the-art large-scale transformer video model UniFormerV2~\cite{a85}. 
UniFormerV2 focuses only on the visual modality and neglecting the audio modality. 
However, we perform fully supervised training by leveraging both visual and audio modalities to establish the upper bound of fully supervised performance. 
To ensure a fair comparison with UniFormerV2, we unify the modality  for our experiments. 
Specifically, we test UniFormerV2 on the AVK-400 dataset. 
Due to limited computational resources, we focus on the model UniFormerV2-L/14 with Frame $16\times3\times4$, we also implement our method on the same dataset, considering solely the visual modality (HAD-V) and both modalities (HAD). The results are presented in Table~\ref{compared_with_LSFSM}. 
From Table~\ref{compared_with_LSFSM}, it is evident that UniFormerV2, utilizing large-scale transformers as their backbone, significantly outperform our HAD-V method on the AVK-400 dataset, with improvements of 17.0\%, respectively. 
Furthermore, our HAD considering both audio and visual modalities surpasses HAD-V focusing only on visual information. However, the performance of HAD on AVK-400 still falls short of UniFormerV2 with massive model parameters, by 8.3\%, respectively. These findings affirm the validity of the upper bound of joint-training performance in Table~\ref{main-results-tab} and demonstrate the benefit of integrating both audio and visual modalities.
This also inspires us to focus on developing multi-modal large-scale transformer video models in the future.

\begin{table}
\centering
\scriptsize
\setlength{\tabcolsep}{14.5pt}
\footnotesize
\renewcommand{\arraystretch}{1.2}
\caption{Comparison of fully-supervised learning in AVK-400.}
\label{compared_with_LSFSM}

{
\begin{tabular}{c|c c|c}
\shline
Methods & Visual & Audio & AVK-400 \\
\hline
UniFormerV2~\cite{a85} & \cmark & \xmark & 93.1 \\
\hline
HAD-V (Ours) & \cmark & \xmark & 76.1 \\
\hline
HAD (Ours) & \cmark & \cmark & 84.8 \\
\shline
\end{tabular}
}
\end{table}

\subsubsection{Analysis of memory usage } 
Many existing class-incremental learning methods store images or frames of historical categories to preserve past knowledge. However, our method stores features of historical categories, thereby saving historical knowledge. We compare the storage of images/frames and features of historical categories to demonstrate the effectiveness of storing features, shown in Table~\ref{Analysis_MU}.
In assessing the memory usage for video processing and feature storage, consider a 10-second video at 8 frames per second, with each frame measuring $3\times224\times224$ pixels and an 8-bit color depth per channel. 
A single frame requires roughly 150,528 bytes, or about 0.143 MB. 
For the whole clip, the total memory is approximately 12,042,240 bytes, or 11.49 MB. For feature storage, where each feature is a 32-bit float (4 bytes), the requirements are as follows: audio features ($10 \times 128$) use 5,120 bytes, 2D visual features ($80 \times 2048$) require 655,360 bytes, and 3D visual features ($10 \times 512$) need 20,480 bytes. 
The total memory for all features per video is thus around 680,960 bytes, or 0.65 MB. 
In the context of the AVK-400 continuous learning task with 4,000 videos, storing frames would need about 44.87 GB, while storing features would only need around 2.54 GB. This demonstrates a clear benefit of feature storage over raw frame data in terms of memory efficiency. Feature storage occupies significantly less space, just a fraction of what's required for raw frames. This efficiency is particularly valuable in large-scale machine learning projects, where optimizing data storage and processing is key.

\begin{table}
  \centering
  \scriptsize
  \setlength{\tabcolsep}{8.5pt}
  \footnotesize
  \renewcommand{\arraystretch}{1.2}
  \caption{Analyzing different data storage methods in terms of memory usage.}
  \label{Analysis_MU}
  {
    \begin{tabular}{c|c|c}
      \shline
      Data Storage Methods & Per Video & Exemplar Data in AVK-400 \\
      \hline
      Frames & 11.49 MB & 44.87 GB \\
      \hline
      Features & 0.65 MB & 2.54 GB \\
      \shline
    \end{tabular}
  }
\end{table}

\section{Conclusion}
This work investigates a fundamental audio-visual problem: Class Incremental Audio-Visual Video Recognition (CIAVVR).
We propose a novel Hierarchical Augmentation and Distillation (HAD) framework for CIAVVR, considering the hierarchical structure in model and video data.
The evaluations of four benchmarks confirm the effectiveness of the proposed HAD.
In the future, we will explore how to use the hierarchical structure in video data to store knowledge of old classes in a non-exemplar way.


%





\ifCLASSOPTIONcaptionsoff
  \newpage
\fi



\bibliographystyle{IEEEtran}
\bibliography{IEEEabrv,egbib}

\begin{thebibliography}{10}
\providecommand{\url}[1]{#1}
\csname url@samestyle\endcsname
\providecommand{\newblock}{\relax}
\providecommand{\bibinfo}[2]{#2}
\providecommand{\BIBentrySTDinterwordspacing}{\spaceskip=0pt\relax}
\providecommand{\BIBentryALTinterwordstretchfactor}{4}
\providecommand{\BIBentryALTinterwordspacing}{\spaceskip=\fontdimen2\font plus
\BIBentryALTinterwordstretchfactor\fontdimen3\font minus \fontdimen4\font\relax}
\providecommand{\BIBforeignlanguage}[2]{{%
\expandafter\ifx\csname l@#1\endcsname\relax
\typeout{** WARNING: IEEEtran.bst: No hyphenation pattern has been}%
\typeout{** loaded for the language `#1'. Using the pattern for}%
\typeout{** the default language instead.}%
\else
\language=\csname l@#1\endcsname
\fi
#2}}
\providecommand{\BIBdecl}{\relax}
\BIBdecl

\bibitem{a8}
Y.~Tian, D.~Li, and C.~Xu, ``Unified multisensory perception: Weakly-supervised audio-visual video parsing,'' in \emph{European Conference on Computer Vision}.\hskip 1em plus 0.5em minus 0.4em\relax Springer, 2020, pp. 436--454.

\bibitem{a9}
Y.~Chen, Y.~Xian, A.~Koepke, Y.~Shan, and Z.~Akata, ``Distilling audio-visual knowledge by compositional contrastive learning,'' in \emph{Proceedings of the IEEE/CVF Conference on Computer Vision and Pattern Recognition}, 2021, pp. 7016--7025.

\bibitem{a71}
Y.~Ban, X.~Alameda-Pineda, L.~Girin, and R.~Horaud, ``Variational bayesian inference for audio-visual tracking of multiple speakers,'' \emph{IEEE Transactions on Pattern Analysis and Machine Intelligence}, vol.~43, no.~5, pp. 1761--1776, 2021.

\bibitem{a72}
I.~D. Gebru, X.~Alameda-Pineda, F.~Forbes, and R.~Horaud, ``Em algorithms for weighted-data clustering with application to audio-visual scene analysis,'' \emph{IEEE Transactions on Pattern Analysis and Machine Intelligence}, vol.~38, no.~12, pp. 2402--2415, 2016.

\bibitem{a73}
T.~Afouras, J.~S. Chung, A.~Senior, O.~Vinyals, and A.~Zisserman, ``Deep audio-visual speech recognition,'' \emph{IEEE Transactions on Pattern Analysis and Machine Intelligence}, pp. 1--1, 2018.

\bibitem{a3}
A.~Krizhevsky, I.~Sutskever, and G.~E. Hinton, ``Imagenet classification with deep convolutional neural networks,'' \emph{Advances in neural information processing systems}, vol.~25, 2012.

\bibitem{a4}
K.~He, X.~Zhang, S.~Ren, and J.~Sun, ``Deep residual learning for image recognition,'' in \emph{Proceedings of the IEEE conference on computer vision and pattern recognition}, 2016, pp. 770--778.

\bibitem{a1}
M.~McCloskey and N.~J. Cohen, ``Catastrophic interference in connectionist networks: The sequential learning problem,'' in \emph{Psychology of learning and motivation}.\hskip 1em plus 0.5em minus 0.4em\relax Elsevier, 1989, vol.~24, pp. 109--165.

\bibitem{a2}
G.~I. Parisi, R.~Kemker, J.~L. Part, C.~Kanan, and S.~Wermter, ``Continual lifelong learning with neural networks: A review,'' \emph{Neural Networks}, vol. 113, pp. 54--71, 2019.

\bibitem{a11}
S.-A. Rebuffi, A.~Kolesnikov, G.~Sperl, and C.~H. Lampert, ``icarl: Incremental classifier and representation learning,'' in \emph{Proceedings of the IEEE conference on Computer Vision and Pattern Recognition}, 2017, pp. 2001--2010.

\bibitem{a6}
E.~Belouadah and A.~Popescu, ``Il2m: Class incremental learning with dual memory,'' in \emph{Proceedings of the IEEE/CVF International Conference on Computer Vision}, 2019, pp. 583--592.

\bibitem{a14}
F.~Zhu, Z.~Cheng, X.-y. Zhang, and C.-l. Liu, ``Class-incremental learning via dual augmentation,'' \emph{Advances in Neural Information Processing Systems}, vol.~34, 2021.

\bibitem{a13}
S.~Mittal, S.~Galesso, and T.~Brox, ``Essentials for class incremental learning,'' in \emph{Proceedings of the IEEE/CVF Conference on Computer Vision and Pattern Recognition}, 2021, pp. 3513--3522.

\bibitem{a16}
H.~Cha, J.~Lee, and J.~Shin, ``Co2l: Contrastive continual learning,'' in \emph{Proceedings of the IEEE/CVF International Conference on Computer Vision}, 2021, pp. 9516--9525.

\bibitem{a74}
C.~V. Nguyen, Y.~Li, T.~D. Bui, and R.~E. Turner, ``Variational continual learning,'' in \emph{International Conference on Learning Representations}, 2018.

\bibitem{a75}
A.~Kumar, S.~Chatterjee, and P.~Rai, ``Bayesian structural adaptation for continual learning,'' in \emph{International Conference on Machine Learning}.\hskip 1em plus 0.5em minus 0.4em\relax PMLR, 2021, pp. 5850--5860.

\bibitem{a76}
M.~M. Derakhshani, X.~Zhen, L.~Shao, and C.~Snoek, ``Kernel continual learning,'' in \emph{International Conference on Machine Learning}.\hskip 1em plus 0.5em minus 0.4em\relax PMLR, 2021, pp. 2621--2631.

\bibitem{a77}
M.~K. Titsias, J.~Schwarz, A.~G. d.~G. Matthews, R.~Pascanu, and Y.~W. Teh, ``Functional regularisation for continual learning with gaussian processes,'' in \emph{International Conference on Learning Representations}, 2019.

\bibitem{a78}
P.~Pan, S.~Swaroop, A.~Immer, R.~Eschenhagen, R.~Turner, and M.~E.~E. Khan, ``Continual deep learning by functional regularisation of memorable past,'' \emph{Advances in Neural Information Processing Systems}, vol.~33, pp. 4453--4464, 2020.

\bibitem{a17}
J.~Kirkpatrick, R.~Pascanu, N.~Rabinowitz, J.~Veness, G.~Desjardins, A.~A. Rusu, K.~Milan, J.~Quan, T.~Ramalho, A.~Grabska-Barwinska \emph{et~al.}, ``Overcoming catastrophic forgetting in neural networks,'' \emph{Proceedings of the national academy of sciences}, vol. 114, no.~13, pp. 3521--3526, 2017.

\bibitem{a18}
F.~Zenke, B.~Poole, and S.~Ganguli, ``Continual learning through synaptic intelligence,'' in \emph{International Conference on Machine Learning}.\hskip 1em plus 0.5em minus 0.4em\relax PMLR, 2017, pp. 3987--3995.

\bibitem{a19}
R.~Aljundi, F.~Babiloni, M.~Elhoseiny, M.~Rohrbach, and T.~Tuytelaars, ``Memory aware synapses: Learning what (not) to forget,'' in \emph{Proceedings of the European Conference on Computer Vision (ECCV)}, 2018, pp. 139--154.

\bibitem{a58}
A.~Chaudhry, P.~K. Dokania, T.~Ajanthan, and P.~H.~S. Torr, ``Riemannian walk for incremental learning: Understanding forgetting and intransigence,'' in \emph{Computer Vision -- ECCV 2018}, V.~Ferrari, M.~Hebert, C.~Sminchisescu, and Y.~Weiss, Eds.\hskip 1em plus 0.5em minus 0.4em\relax Cham: Springer International Publishing, 2018, pp. 556--572.

\bibitem{a59}
K.~Joseph, S.~Khan, F.~S. Khan, R.~M. Anwer, and V.~N. Balasubramanian, ``Energy-based latent aligner for incremental learning,'' in \emph{Proceedings of the IEEE/CVF Conference on Computer Vision and Pattern Recognition}, 2022, pp. 7452--7461.

\bibitem{a20}
A.~A. Rusu, N.~C. Rabinowitz, G.~Desjardins, H.~Soyer, J.~Kirkpatrick, K.~Kavukcuoglu, R.~Pascanu, and R.~Hadsell, ``Progressive neural networks,'' \emph{arXiv preprint arXiv:1606.04671}, 2016.

\bibitem{a21}
J.~Schwarz, W.~Czarnecki, J.~Luketina, A.~Grabska-Barwinska, Y.~W. Teh, R.~Pascanu, and R.~Hadsell, ``Progress \& compress: A scalable framework for continual learning,'' in \emph{International Conference on Machine Learning}.\hskip 1em plus 0.5em minus 0.4em\relax PMLR, 2018, pp. 4528--4537.

\bibitem{a22}
X.~Li, Y.~Zhou, T.~Wu, R.~Socher, and C.~Xiong, ``Learn to grow: A continual structure learning framework for overcoming catastrophic forgetting,'' in \emph{International Conference on Machine Learning}.\hskip 1em plus 0.5em minus 0.4em\relax PMLR, 2019, pp. 3925--3934.

\bibitem{a23}
Z.~Wu, C.~Baek, C.~You, and Y.~Ma, ``Incremental learning via rate reduction,'' in \emph{Proceedings of the IEEE/CVF Conference on Computer Vision and Pattern Recognition}, 2021, pp. 1125--1133.

\bibitem{a54}
A.~Mallya and S.~Lazebnik, ``Packnet: Adding multiple tasks to a single network by iterative pruning,'' in \emph{2018 IEEE/CVF Conference on Computer Vision and Pattern Recognition}, 2018, pp. 7765--7773.

\bibitem{a55}
S.~Ebrahimi, F.~Meier, R.~Calandra, T.~Darrell, and M.~Rohrbach, ``Adversarial continual learning,'' \emph{ArXiv}, vol. abs/2003.09553, 2020.

\bibitem{a56}
S.~Yan, J.~Xie, and X.~He, ``Der: Dynamically expandable representation for class incremental learning,'' in \emph{2021 IEEE/CVF Conference on Computer Vision and Pattern Recognition (CVPR)}, 2021, pp. 3013--3022.

\bibitem{a57}
F.-Y. Wang, D.-W. Zhou, H.-J. Ye, and D.-C. Zhan, ``Foster: Feature boosting and compression for class-incremental learning,'' \emph{ECCV 2022}, 2022.

\bibitem{a27}
R.~Kemker and C.~Kanan, ``Fearnet: Brain-inspired model for incremental learning,'' \emph{arXiv preprint arXiv:1711.10563}, 2017.

\bibitem{a69}
M.~Boschini, L.~Bonicelli, P.~Buzzega, A.~Porrello, and S.~Calderara, ``Class-incremental continual learning into the extended der-verse,'' \emph{IEEE Transactions on Pattern Analysis and Machine Intelligence}, pp. 1--16, 2022.

\bibitem{a12}
S.~Hou, X.~Pan, C.~C. Loy, Z.~Wang, and D.~Lin, ``Learning a unified classifier incrementally via rebalancing,'' in \emph{Proceedings of the IEEE/CVF Conference on Computer Vision and Pattern Recognition}, 2019, pp. 831--839.

\bibitem{a24}
H.~Shin, J.~K. Lee, J.~Kim, and J.~Kim, ``Continual learning with deep generative replay,'' \emph{Advances in neural information processing systems}, vol.~30, 2017.

\bibitem{a25}
C.~Wu, L.~Herranz, X.~Liu, J.~van~de Weijer, B.~Raducanu \emph{et~al.}, ``Memory replay gans: Learning to generate new categories without forgetting,'' \emph{Advances in Neural Information Processing Systems}, vol.~31, 2018.

\bibitem{a26}
L.~Wang, K.~Yang, C.~Li, L.~Hong, Z.~Li, and J.~Zhu, ``Ordisco: Effective and efficient usage of incremental unlabeled data for semi-supervised continual learning,'' in \emph{Proceedings of the IEEE/CVF Conference on Computer Vision and Pattern Recognition}, 2021, pp. 5383--5392.

\bibitem{a60}
A.~Douillard, M.~Cord, C.~Ollion, T.~Robert, and E.~Valle, ``Podnet: Pooled outputs distillation for small-tasks incremental learning,'' in \emph{European Conference on Computer Vision}.\hskip 1em plus 0.5em minus 0.4em\relax Springer, 2020, pp. 86--102.

\bibitem{a61}
X.~Tao, X.~Chang, X.~Hong, X.~Wei, and Y.~Gong, ``Topology-preserving class-incremental learning,'' in \emph{European Conference on Computer Vision}.\hskip 1em plus 0.5em minus 0.4em\relax Springer, 2020, pp. 254--270.

\bibitem{a63}
F.~Zhu, X.-Y. Zhang, C.~Wang, F.~Yin, and C.-L. Liu, ``Prototype augmentation and self-supervision for incremental learning,'' in \emph{Proceedings of the IEEE/CVF Conference on Computer Vision and Pattern Recognition}, 2021, pp. 5871--5880.

\bibitem{a62}
A.~Ashok, K.~Joseph, and V.~Balasubramanian, ``Class-incremental learning with cross-space clustering and controlled transfer,'' \emph{ECCV 2022}, 2022.

\bibitem{a80}
A.~Villa, K.~Alhamoud, V.~Escorcia, F.~Caba, J.~L. Alc\'azar, and B.~Ghanem, ``vclimb: A novel video class incremental learning benchmark,'' in \emph{Proceedings of the IEEE/CVF Conference on Computer Vision and Pattern Recognition (CVPR)}, June 2022, pp. 19\,035--19\,044.

\bibitem{a82}
\BIBentryALTinterwordspacing
Y.~Pei, Z.~Qing, J.~CEN, X.~Wang, S.~Zhang, Y.~Wang, M.~Tang, N.~Sang, and X.~Qian, ``Learning a condensed frame for memory-efficient video class-incremental learning,'' in \emph{Advances in Neural Information Processing Systems}, A.~H. Oh, A.~Agarwal, D.~Belgrave, and K.~Cho, Eds., 2022. [Online]. Available: \url{https://openreview.net/forum?id=lCGYC7pXWNQ}
\BIBentrySTDinterwordspacing

\bibitem{a64}
Y.~Tan, Y.~Hao, X.~He, Y.~Wei, and X.~Yang, ``Selective dependency aggregation for action classification,'' in \emph{Proceedings of the 29th ACM International Conference on Multimedia}, 2021, pp. 592--601.

\bibitem{a65}
R.~Girdhar, D.~Ramanan, A.~Gupta, J.~Sivic, and B.~Russell, ``Actionvlad: Learning spatio-temporal aggregation for action classification,'' in \emph{Proceedings of the IEEE conference on computer vision and pattern recognition}, 2017, pp. 971--980.

\bibitem{a66}
L.~Wang, W.~Li, W.~Li, and L.~Van~Gool, ``Appearance-and-relation networks for video classification,'' in \emph{Proceedings of the IEEE conference on computer vision and pattern recognition}, 2018, pp. 1430--1439.

\bibitem{a67}
S.~Savarese, A.~DelPozo, J.~C. Niebles, and L.~Fei-Fei, ``Spatial-temporal correlatons for unsupervised action classification,'' in \emph{2008 IEEE Workshop on Motion and video Computing}.\hskip 1em plus 0.5em minus 0.4em\relax IEEE, 2008, pp. 1--8.

\bibitem{a35}
Y.~Aytar, C.~Vondrick, and A.~Torralba, ``See, hear, and read: Deep aligned representations,'' \emph{arXiv preprint arXiv:1706.00932}, 2017.

\bibitem{a36}
V.~Sanguineti, P.~Morerio, N.~Pozzetti, D.~Greco, M.~Cristani, and V.~Murino, ``Leveraging acoustic images for effective self-supervised audio representation learning,'' in \emph{European Conference on Computer Vision}.\hskip 1em plus 0.5em minus 0.4em\relax Springer, 2020, pp. 119--135.

\bibitem{a37}
S.~Ma, Z.~Zeng, D.~McDuff, and Y.~Song, ``Active contrastive learning of audio-visual video representations,'' \emph{arXiv preprint arXiv:2009.09805}, 2020.

\bibitem{a38}
S.~Lee, J.~Chung, Y.~Yu, G.~Kim, T.~Breuel, G.~Chechik, and Y.~Song, ``Acav100m: Automatic curation of large-scale datasets for audio-visual video representation learning,'' in \emph{Proceedings of the IEEE/CVF International Conference on Computer Vision}, 2021, pp. 10\,274--10\,284.

\bibitem{a53}
A.~Zheng, M.~Hu, B.~Jiang, Y.~Huang, Y.~Yan, and B.~Luo, ``Adversarial-metric learning for audio-visual cross-modal matching,'' \emph{IEEE Transactions on Multimedia}, vol.~24, pp. 338--351, 2022.

\bibitem{a28}
Y.~Tian, J.~Shi, B.~Li, Z.~Duan, and C.~Xu, ``Audio-visual event localization in unconstrained videos,'' in \emph{Proceedings of the European Conference on Computer Vision (ECCV)}, 2018, pp. 247--263.

\bibitem{a29}
Y.~Tian, D.~Li, and C.~Xu, ``Unified multisensory perception: Weakly-supervised audio-visual video parsing,'' in \emph{European Conference on Computer Vision}.\hskip 1em plus 0.5em minus 0.4em\relax Springer, 2020, pp. 436--454.

\bibitem{a30}
Y.~Wu, L.~Zhu, Y.~Yan, and Y.~Yang, ``Dual attention matching for audio-visual event localization,'' in \emph{Proceedings of the IEEE/CVF international conference on computer vision}, 2019, pp. 6292--6300.

\bibitem{a31}
Y.-B. Lin, H.-Y. Tseng, H.-Y. Lee, Y.-Y. Lin, and M.-H. Yang, ``Exploring cross-video and cross-modality signals for weakly-supervised audio-visual video parsing,'' \emph{Advances in Neural Information Processing Systems}, vol.~34, 2021.

\bibitem{a51}
S.~Liu, W.~Quan, C.~Wang, Y.~Liu, B.~Liu, and D.-M. Yan, ``Dense modality interaction network for audio-visual event localization,'' \emph{IEEE Transactions on Multimedia}, pp. 1--1, 2022.

\bibitem{a52}
C.~Xue, X.~Zhong, M.~Cai, H.~Chen, and W.~Wang, ``Audio-visual event localization by learning spatial and semantic co-attention,'' \emph{IEEE Transactions on Multimedia}, pp. 1--1, 2021.

\bibitem{a32}
R.~Gao, R.~Feris, and K.~Grauman, ``Learning to separate object sounds by watching unlabeled video,'' in \emph{Proceedings of the European Conference on Computer Vision (ECCV)}, 2018, pp. 35--53.

\bibitem{a33}
C.~Gan, D.~Huang, H.~Zhao, J.~B. Tenenbaum, and A.~Torralba, ``Music gesture for visual sound separation,'' in \emph{Proceedings of the IEEE/CVF Conference on Computer Vision and Pattern Recognition}, 2020, pp. 10\,478--10\,487.

\bibitem{a34}
Y.~Tian, D.~Hu, and C.~Xu, ``Cyclic co-learning of sounding object visual grounding and sound separation,'' in \emph{Proceedings of the IEEE/CVF Conference on Computer Vision and Pattern Recognition}, 2021, pp. 2745--2754.

\bibitem{a42}
X.~Wang, Y.-F. Wang, and W.~Y. Wang, ``Watch, listen, and describe: Globally and locally aligned cross-modal attentions for video captioning,'' \emph{arXiv preprint arXiv:1804.05448}, 2018.

\bibitem{a43}
T.~Rahman, B.~Xu, and L.~Sigal, ``Watch, listen and tell: Multi-modal weakly supervised dense event captioning,'' in \emph{Proceedings of the IEEE/CVF International Conference on Computer Vision}, 2019, pp. 8908--8917.

\bibitem{a44}
Y.~Tian, C.~Guan, J.~Goodman, M.~Moore, and C.~Xu, ``Audio-visual interpretable and controllable video captioning,'' in \emph{IEEE Computer Society Conference on Computer Vision and Pattern Recognition workshops}, 2019.

\bibitem{a39}
M.~Subedar, R.~Krishnan, P.~L. Meyer, O.~Tickoo, and J.~Huang, ``Uncertainty-aware audiovisual activity recognition using deep bayesian variational inference,'' in \emph{Proceedings of the IEEE/CVF international conference on computer vision}, 2019, pp. 6301--6310.

\bibitem{a40}
E.~Kazakos, A.~Nagrani, A.~Zisserman, and D.~Damen, ``Epic-fusion: Audio-visual temporal binding for egocentric action recognition,'' in \emph{Proceedings of the IEEE/CVF International Conference on Computer Vision}, 2019, pp. 5492--5501.

\bibitem{a41}
A.~Nagrani, C.~Sun, D.~Ross, R.~Sukthankar, C.~Schmid, and A.~Zisserman, ``Speech2action: Cross-modal supervision for action recognition,'' in \emph{Proceedings of the IEEE/CVF Conference on Computer Vision and Pattern Recognition}, 2020, pp. 10\,317--10\,326.

\bibitem{a48}
\BIBentryALTinterwordspacing
M.~O'Searcoid, \emph{Metric Spaces}, ser. Springer Undergraduate Mathematics Series.\hskip 1em plus 0.5em minus 0.4em\relax Springer London, 2006. [Online]. Available: \url{https://books.google.com.hk/books?id=aP37I4QWFRcC}
\BIBentrySTDinterwordspacing

\bibitem{a49}
H.~Kim, G.~Papamakarios, and A.~Mnih, ``The lipschitz constant of self-attention,'' in \emph{International Conference on Machine Learning}.\hskip 1em plus 0.5em minus 0.4em\relax PMLR, 2021, pp. 5562--5571.

\bibitem{a45}
J.~F. Gemmeke, D.~P. Ellis, D.~Freedman, A.~Jansen, W.~Lawrence, R.~C. Moore, M.~Plakal, and M.~Ritter, ``Audio set: An ontology and human-labeled dataset for audio events,'' in \emph{2017 IEEE international conference on acoustics, speech and signal processing (ICASSP)}.\hskip 1em plus 0.5em minus 0.4em\relax IEEE, 2017, pp. 776--780.

\bibitem{kay2017kinetics}
W.~Kay, J.~Carreira, K.~Simonyan, B.~Zhang, C.~Hillier, S.~Vijayanarasimhan, F.~Viola, T.~Green, T.~Back, P.~Natsev \emph{et~al.}, ``The kinetics human action video dataset,'' \emph{arXiv preprint arXiv:1705.06950}, 2017.

\bibitem{a68}
P.~Buzzega, M.~Boschini, A.~Porrello, D.~Abati, and S.~Calderara, ``Dark experience for general continual learning: a strong, simple baseline,'' in \emph{Advances in Neural Information Processing Systems}, H.~Larochelle, M.~Ranzato, R.~Hadsell, M.~F. Balcan, and H.~Lin, Eds., vol.~33.\hskip 1em plus 0.5em minus 0.4em\relax Curran Associates, Inc., 2020, pp. 15\,920--15\,930.

\bibitem{a86}
W.~Shi and M.~Ye, ``Prototype reminiscence and augmented asymmetric knowledge aggregation for non-exemplar class-incremental learning,'' in \emph{Proceedings of the IEEE/CVF International Conference on Computer Vision}, 2023, pp. 1772--1781.

\bibitem{a46}
D.~Tran, H.~Wang, L.~Torresani, J.~Ray, Y.~LeCun, and M.~Paluri, ``A closer look at spatiotemporal convolutions for action recognition,'' in \emph{Proceedings of the IEEE conference on Computer Vision and Pattern Recognition}, 2018, pp. 6450--6459.

\bibitem{5206848}
J.~Deng, W.~Dong, R.~Socher, L.-J. Li, K.~Li, and L.~Fei-Fei, ``Imagenet: A large-scale hierarchical image database,'' in \emph{2009 IEEE Conference on Computer Vision and Pattern Recognition}, 2009, pp. 248--255.

\bibitem{7952261}
J.~F. Gemmeke, D.~P.~W. Ellis, D.~Freedman, A.~Jansen, W.~Lawrence, R.~C. Moore, M.~Plakal, and M.~Ritter, ``Audio set: An ontology and human-labeled dataset for audio events,'' in \emph{2017 IEEE International Conference on Acoustics, Speech and Signal Processing (ICASSP)}, 2017, pp. 776--780.

\bibitem{a47}
D.~P. Kingma and J.~Ba, ``Adam: A method for stochastic optimization,'' \emph{arXiv preprint arXiv:1412.6980}, 2014.

\bibitem{a85}
K.~Li, Y.~Wang, Y.~He, Y.~Li, Y.~Wang, L.~Wang, and Y.~Qiao, ``Uniformerv2: Spatiotemporal learning by arming image vits with video uniformer,'' \emph{arXiv preprint arXiv:2211.09552}, 2022.

\end{thebibliography}

\begin{IEEEbiography}[{\includegraphics[width=1in,height=1.25in,clip,keepaspectratio]{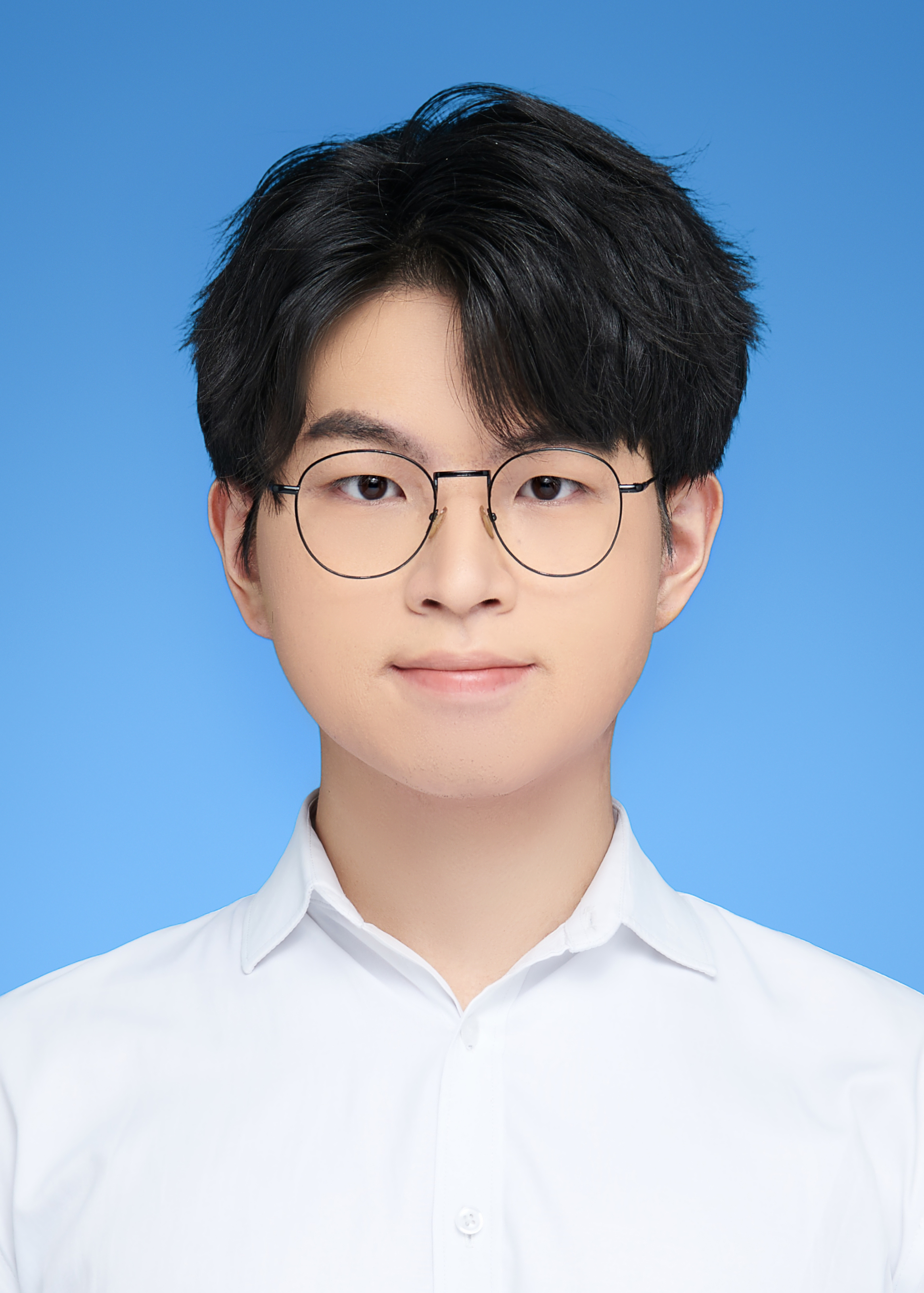}}]{Yukun Zuo}(Member, IEEE) received the B.S. degree in information security from the University of Science and Technology of China in 2018. He received the Ph.D. degree in Information and Communication Engineering from the University of Science and Technology of China in 2023. He is currently a Research Fellow in the Department of Civil and Environmental Engineering at the University of Michigan. His research interests include computer vision and machine learning.
\end{IEEEbiography}

\begin{IEEEbiography}[{\includegraphics[width=1in,height=1.25in,clip,keepaspectratio]{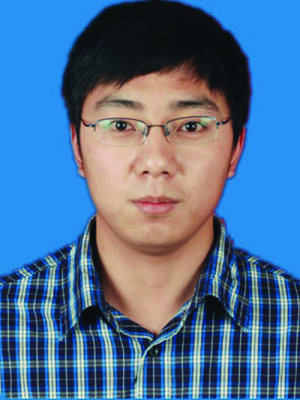}}]{Hantao Yao}(Member, IEEE) received the B.S. degree from XiDian University, Xi’an, China, in 2012. He received his Ph.D. degree in Institute of Computing Technology, University of Chinese Academy of Sciences in 2018. After graduation, he worked as a post-doctoral from 2018 to 2020 at National Laboratory of Pattern Recognition, Institute of Automation, Chinese Academy of Sciences. Now, he is an associate professor at State Key Laboratory of Multimodal Artificial Intelligence Systems, Institute of Automation, Chinese Academy of Sciences. He is the recipient of National Postdoctoral Programme for Innovative Talents. His current research interests are zero-shot learning, person tracking and detection, person re-identification.
\end{IEEEbiography}

\begin{IEEEbiography}[{\includegraphics[width=1in,height=1.25in,clip,keepaspectratio]{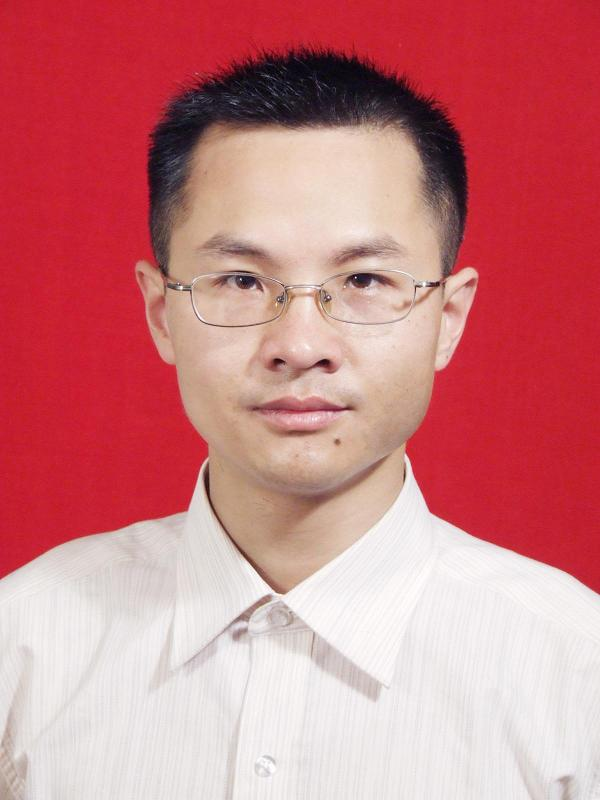}}]{Liansheng Zhuang}(Member, IEEE) received the bachelor's and Ph.D. degrees from the University of Science and Technology of China (USTC), China, in 2001 and 2006, respectively. In 2011, he was nominated to join the STARTRACKER Project of Microsoft Research of Asia (MSRA), and he was a Vendor Researcher with the Visual Computing Group, Microsoft Research, Beijing. From 2012 to 2013, he was a Visiting Research Scientist with the Department of EECS, University of California at Berkeley, Berkeley. He is currently an Associate Professor with the School of Information Science and Technology, USTC. His main research interesting is in computer vision, and machine learning. He is a member of ACM, IEEE and CCF.
\end{IEEEbiography}

\begin{IEEEbiography}[{\includegraphics[width=1in,height=1.25in,clip,keepaspectratio]{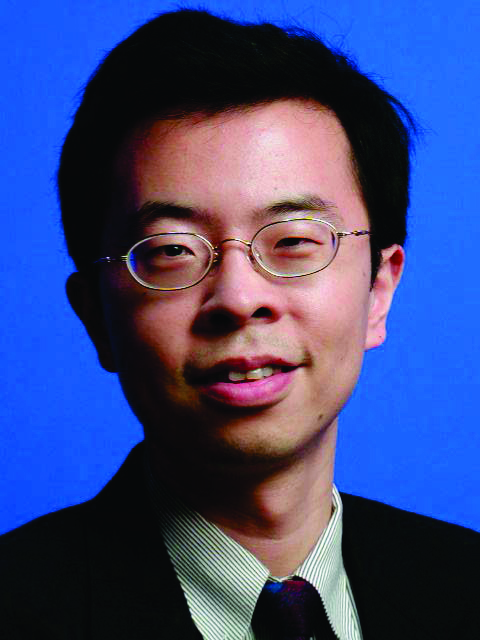}}]{Changsheng Xu}(Fellow, IEEE) is a Professor in State Key Laboratory of Multimodal Artificial Intelligence Systems, Institute of Automation, Chinese Academy of Sciences and Executive Director of China-Singapore Institute of Digital Media. His research interests include multimedia content analysis/indexing/retrieval, pattern recognition and computer vision. He has hold 50 granted/pending patents and published over 400 refereed research papers in these areas. Dr.Xu is an Associate Editor of IEEE Trans. on Multimedia, ACM Trans. on Multimedia Computing, Communications and Applications and ACM/Springer Multimedia Systems Journal. He received the Best Associate Editor Award of ACM Trans. on Multimedia Computing, Communications and Applications in 2012 and the Best Editorial Member Award of ACM/Springer Multimedia Systems Journal in 2008. He served as Program Chair of ACM Multimedia 2009. He has served as associate editor, guest editor, general chair, program chair, area/track chair, special session organizer, session chair and TPC member for over 20 IEEE and ACM prestigious multimedia journals, conferences and workshops. He is IEEE Fellow, IAPR Fellow and ACM Distinguished Scientist.\\
\end{IEEEbiography}

\end{document}